\pdfoutput=1

\documentclass[11pt]{article}

\usepackage{EMNLP2022}

\usepackage{times}
\usepackage{latexsym}

\usepackage[T1]{fontenc}

\usepackage[utf8]{inputenc}

\usepackage{microtype}

\usepackage{inconsolata}

\usepackage{url}
\usepackage{amsmath}
\usepackage{amssymb}
\usepackage{booktabs}
\usepackage{graphicx}
\usepackage{varwidth}
\usepackage{tabularx}
\usepackage{algorithm}
\usepackage[noend]{algpseudocode}

\usepackage[utf8]{inputenc}

\usepackage[textsize=scriptsize,disable]{todonotes}

\title{Natural Language Deduction with Incomplete Information}

\author{Zayne Sprague \quad\quad
  Kaj Bostrom \quad\quad
  Swarat Chaudhuri \quad\quad
  Greg Durrett \\
  Department of Computer Science\\
  The University of Texas at Austin\\
  \texttt{zaynesprague@utexas.edu, \{kaj,swarat,gdurrett\}@cs.utexas.edu}
  }

\date{}

\begin{document}
\maketitle
\begin{abstract}
A growing body of work studies how to answer a question or verify a claim by generating a natural language ``proof'': a chain of deductive inferences yielding the answer based on a set of premises. However, these methods can only make sound deductions when they follow from evidence that is given. We propose a new system that can handle the underspecified setting where not all premises are stated at the outset; that is, additional assumptions need to be materialized to prove a claim. By using a natural language generation model to abductively infer a premise given another premise and a conclusion, we can impute missing pieces of evidence needed for the conclusion to be true. Our system searches over two fringes in a bidirectional fashion, interleaving deductive (forward-chaining)
and abductive (backward-chaining) generation steps. We sample multiple possible outputs for each step to achieve coverage of the search space, at the same time ensuring correctness by filtering low-quality generations with a round-trip validation procedure. Results on a modified version of the EntailmentBank dataset and a new dataset called \emph{Everyday Norms: Why Not?} show that abductive generation with validation can recover premises across in- and out-of-domain settings.\footnote{Code and data publicly available at \url{https://github.com/Zayne-sprague/Natural_Language_Deduction_with_Incomplete_Information.git}}
\end{abstract}

\section{Introduction}

Substantial prior work in domains like question answering \cite{rajpurkar-etal-2016-squad, yang-etal-2018-hotpotqa, kwiatkowski-etal-2019-natural}, textual entailment \cite{bowman-etal-2015-large, williams-etal-2018-broad, nie-etal-2020-adversarial}, and other types of reasoning \cite{clarkreasoners, dalvi-etal-2021-explaining} deals with making inferences from stated information, where we draw conclusions and answer questions based on textual context provided directly to a model. However, a growing body of research studies the problem of reasoning given incomplete information, especially for tasks labeled as commonsense reasoning \cite{talmor-etal-2019-commonsenseqa, rajani-etal-2019-explain}. Current approaches in these domains often work through latent reasoning by large language models \cite{Lourie2021UNICORNOR}, with only a few explicitly materializing the missing knowledge \cite{bosselut-etal-2019-comet,Bhagavatula2020Abductive,arabshahi2021conversational,liu-etal-2022-generated,katz-inferring}. However, making knowledge explicit is critical to make reasoning processes \emph{explainable}: it allows users to critique those explanations and allows systems to reuse inferred knowledge across scenarios \cite{Dalvi2022TowardsTR}.

\begin{figure}
    \centering
    \includegraphics[width=\columnwidth,trim=3mm 113mm 30mm 28mm]{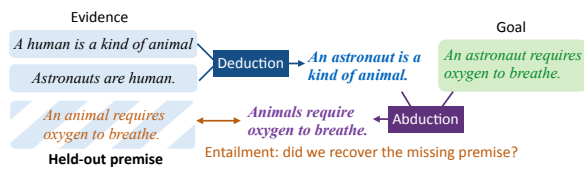}
    \caption{An example of deductive (previous work) and abductive (our work) reasoning used together to search for missing evidence needed to entail a goal in a depth 2 tree from EntailmentBank.}
    \label{fig:abduction_intro}
    \vspace{-2mm}
\end{figure}

\begin{figure*}[t!]
    \centering
    \includegraphics[width=2\columnwidth,trim=3mm 131mm 10mm 28mm]{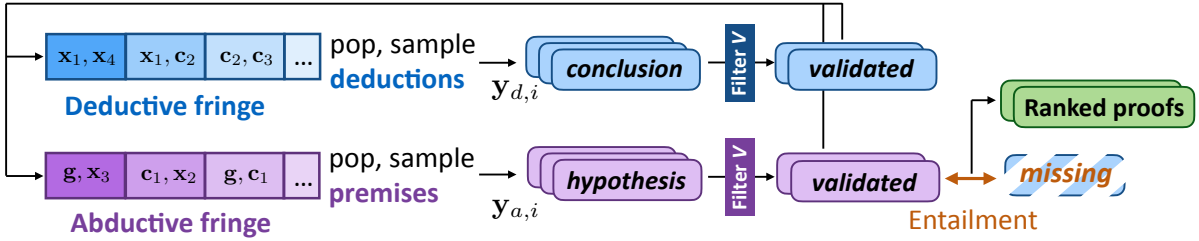}
    \caption{Overview of the ADGV system and its components.  First, on the left, priority queues of possible deductive (blue) and abductive (purple) step inputs give the highest-scoring step inputs for each step type.  Multiple generations are produced at each step, and each sample is validated and discarded if invalid (middle).  Finally (right), the new validated samples are pushed onto the queues following the rules in Table \ref{tab:allowed_steps} and we optionally check whether a particular missing premise has been recovered in our experiments.}
    \label{fig:system_overview}
\end{figure*}

The materialization of new knowledge is naturally formulated as \emph{abduction}: generating an explanation given a premise and a conclusion. Abductive reasoning as a text generation task is fundamentally challenging, as it is 
an underspecified task with a large 
search space of valid generations, hence why prior work has framed it as a multiple-choice problem \cite{Bhagavatula2020Abductive}. Nevertheless, the freeform generation setting is the one that real-world explainable reasoning systems are faced with. 

In this paper, we develop an approach that combines abductive reasoning with multistep deductive reasoning. We build on recent discrete search-based approaches that construct \emph{entailment trees} \cite{dalvi-etal-2021-explaining,bostrom2022natural, yang2021learning, hong2022metgen} to represent deductive inferences in natural language. Although more transparent than discriminative end-to-end models, these methods have so far required all necessary premises to be explicitly provided, and cannot account for abductive reasoning.

Our input is a set of \emph{incomplete} premise facts and a goal; our algorithm searches forward from the premises and backwards from the goal to build a proof that entails the goal \emph{and} recovers a missing premise through a combination of deductive and abductive inferences. Figure \ref{fig:abduction_intro} shows an example. To 
constrain the model's generation, we incorporate a validation criterion to test the consistency of each logical inference. We call this new system ADGV (Abduction and Deductive Generation with Validation, Figure~\ref{fig:system_overview}).  At its core, ADGV follows a similar heuristic search to \citet{bostrom2022natural}, iteratively generating conclusions and adding them to the search frontier, but incorporates abductive steps (analogous to backward-chaining) to make the search two-sided. 

We evaluate on a new task variant that requires recovering a missing premise from a subset of textual evidence and a goal. We use two datasets: EntailmentBank \cite{dalvi-etal-2021-explaining} and \emph{Everyday Norms: Why Not?}, a new dataset that we construct that requires combining information about situations with general principles. 
We assess both \emph{coverage} of held-out premises on our test examples and \emph{step validity} of the steps used to construct them, thereby establishing the ability of ADGV to recover premises as well as construct entailment trees reaching the goal of the original example. Although our approach can reconstruct premises with a high validity rate, achieving high coverage has significant headroom for future work.


Our contributions are: (1) introduction of a new task for natural language understanding, recovering a premise in an underspecified entailment tree, along with a new dataset, \emph{Everyday Norms: Why Not?}; (2) a new abductive step model and ADGV inference method, which combines forward and backward search; (3) new validation techniques that improve step validity in these models.

\section{Problem Description}

We study the task of generating a natural language proof tree $T$ that entails a goal $\mathbf{g}$ given a set of textual evidence $X = \{\mathbf{x}_1\dots \mathbf{x}_n\}$.  Unique to our work, we remove one of the pieces of textual evidence $\mathbf{x}_m$ creating an underspecified setting where a deduction system operating over stated premises \cite{dalvi-etal-2021-explaining,bostrom2022natural} cannot build an entailment tree capable of reaching the goal. The task is then to prove the goal $\mathbf{g}$ while also recovering $\mathbf{x}_m$, which requires searching backwards from the goal to generate missing information. An overview of our abductive reasoning system can be seen in Figure \ref{fig:system_overview}.

Note that there is a trivial solution to this problem, which is to immediately assume that $\mathbf{x}_m = \mathbf{g}$, leading to a vacuous proof. There is no easy way to rule out this solution, as it is hard to come up with a first-order principle for what makes an atomic premise. In existing datasets like EntailmentBank \cite{dalvi-etal-2021-explaining}, premises can be low-level  definitions (``\emph{revolving around means orbiting}'') or more complex process descriptions (``\emph{Photosynthesis means producers / green plants convert from carbon dioxide and water and solar energy into carbohydrates and food and oxygen for themselves}''). Other past work \cite{Dalvi2022TowardsTR, weir2022dynamic} has use large language models to determine atomicity, but this also fails to yield a consistent principle beyond preferring statements that are attested in large web corpora.

As a result, we will use our search procedure to iteratively unroll a goal into simpler statements in an attempt to recover the specific premise $\mathbf{x}_m$ with \emph{some} tree that we find. We will evaluate according to two criteria. Our first criterion is \textbf{recall} of the missing premise at some point along the search process, using a scoring metric $E(\mathbf{x'}_i, \mathbf{x}_m) \in \mathbb{R}$ to determine if a generated premise $\mathbf{x'}_i$ is logically equivalent to $\mathbf{x}_m$. Our second criterion is \textbf{validity} of the tree that yields $\mathbf{x}_m$, judged according to human ratings.

\section{Methods}
\label{sec:methods_intro}

Our approach is based on two generative modules called \emph{step models}. Our deductive step model $S_d$ defines a probability distribution $P_{S_d}(\mathbf{y}\ |\ \mathbf{x}_1\dots \mathbf{x}_n)$ over valid conclusions $\mathbf{y}$ given premises $\mathbf{x}_1\dots \mathbf{x}_n$, all of which are represented as natural language strings. We use the same notion of deduction as in past work
\cite{bostrom-etal-2021-flexible}, where the model should place probability mass over correct and useful conclusions that can be inferred from the premises (i.e., not simply copying a premise). Following past work, we set $n=2$, which we find sufficient to handle both of our datasets.

New in this work, we additionally introduce an \emph{abductive} step model $S_a = P_{S_a}(\mathbf{x'} \mid \mathbf{x}_1\dots \mathbf{x}_n, \mathbf{c})$. This model is meant to ``reverse'' the behavior of the forward model in a similar fashion as backward-chaining in Prolog \cite{robinson1965}. Specifically, it takes a conclusion statement $\mathbf{c}$ as well as one premise $\mathbf{x}$ and generates a hypothesis $\mathbf{x'}$. The generated hypothesis, $\mathbf{x'}$, constitutes a new piece of information that the step model infers is necessary to make the original conclusion $\mathbf{c}$ true.  This operation can then be chained repeatedly to uncover more general and abstract information. We find in our work that setting $n=1$ (one premise and a conclusion $\mathbf{c}$) is sufficient.

\begin{figure}
    \centering
    \includegraphics[width=\columnwidth,trim=3mm 85mm 74mm 28mm]{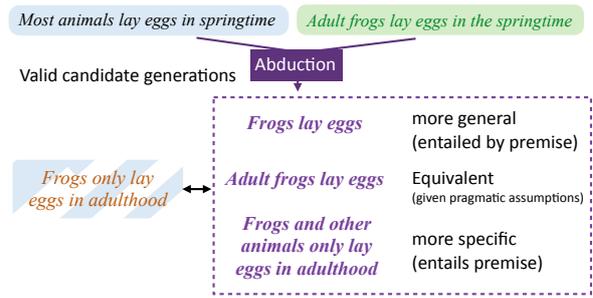}
    \caption{The abductive model can generate numerous valid inferences from a premise (blue) and goal (green) which can relate to the reference premise in a few ways: mutual entailment (middle generation) and entailment in either direction based on whether the generation is more general or more specific than the missing premise.}
    \label{fig:under_and_entail}
\end{figure}

Deductive inferences in the domains we consider may be lexically underspecified, but typically represent a clear logical operation. However, abduction does not.  An example can be seen in Figure \ref{fig:under_and_entail}: the abductive model can produce multiple valid generations at varying levels of specificity. Determining the truth of these generations is extremely challenging as in other work that tries to generate intermediate unstated inferences \cite{rajani-etal-2019-explain,wiegreffe2021reframing,Dalvi2022TowardsTR,liu-etal-2022-generated}. To mitigate this, we introduce round-trip validators which enforce the condition that the forward and abductive models' generations must agree.

\paragraph{Models and Data} Our system revolves around structuring the application of the two step models, $S_d$ and $S_a$, with a search procedure. We first describe the mechanics of the step models and then the heuristic search procedure, which employs two heuristics $H^{d}$ and $H^{a}$ to guide which step generation to perform next.

Both models are trained on data from EntailmentBank \cite{dalvi-etal-2021-explaining}. Following \cite{bostrom2022natural}, we do not rely on complete trees from EntailmentBank to train the step models, but instead view a tree $T$ as a collection of steps $T_i = (\mathbf{x}_{i,1},\ldots,\mathbf{x}_{i,n} \rightarrow \mathbf{c}_i)$.

\subsection{Step Models}

Our \textbf{abductive step model} is an instance of a pre-trained language model. We specialize it to map from a conclusion statement and a single premise to a hypothesized missing premise, yielding the distribution $p_{S_a}(\mathbf{x'} \mid \mathbf{x}, \mathbf{c})$.

The abductive step model is trained on the EntailmentBank dataset by converting each step $T_i = (\mathbf{x}_1,\mathbf{x}_2 \rightarrow \mathbf{c})$ into multiple abductive steps by ablating each input in turn: $\mathbf{x}_1, \mathbf{c} \rightarrow \mathbf{x}_2$ and $\mathbf{x}_2, \mathbf{c} \rightarrow \mathbf{x}_1$.  We ensure the conclusion $\mathbf{c}$ is always appended at the end of the input so the model can learn asymmetric relationships between premises and the input conclusion.
The model is trained with teacher forcing to generate exactly the correct premise; however, during inference we sample as many as $k=40$ generations from the abductive model to account for underspecification.

Our \textbf{deductive step model} follows \citet{bostrom2022natural} and is trained in a similar fashion as the abductive step model. We fine-tune a pretrained language model to map a set of premises to a conclusion statement, giving the distribution $p_{S_d}(\mathbf{c} \mid x_1, ..., x_n)$. This model is trained on the EntailmentBank dataset only (not using data from \citet{bostrom-etal-2021-flexible}), using intermediate steps $T_i\ {=}\ (\mathbf{x}_1,\mathbf{x}_2 \rightarrow \mathbf{c})$ as training examples.  During inference, we sample as many as $k'\ {=}\ 10$ generations from the deductive model to account for underspecification.

\subsection{Search}


\begin{algorithm}[t!]
\small
\caption{Abductive and Deductive Generation with Validation}
\begin{algorithmic}
\State {\hspace{-0.125in}Inputs: a collection of premises $X$, a goal $\mathbf{g}$, and $\mathrm{maxSteps}$}
\State {\hspace{-0.125in}\textbf{procedure} $\textsc{ADGV}(X = \{\mathbf{x}_1\dots \mathbf{x}_n\},\ \mathbf{g}, \mathrm{maxSteps})$:}
\State $fringe_d \gets \{(\mathbf{x}_i, \mathbf{x}_j) |\ \mathbf{x}_i, \mathbf{x}_j \in X, i \neq j \}$
\State $fringe_a \gets \mathrm{pairs}(\mathbf{g}, X)$
\State $seen_d \gets X$
\State $seen_a \gets \{\mathbf{g}\}$
\State $i \gets 1$
\While{$|fringe_d| + |fringe_a| > 0 \wedge i \leq \mathrm{maxSteps}$}
    \State $i \gets i + 1$
    \State $(\mathbf{x}_{d,1}, \mathbf{x}_{d,2}) \leftarrow \text{pop}\  \mathrm{argmax}_{H_d}(fringe_d)$
    \State $(\mathbf{c}_a, \mathbf{x}_a) \leftarrow \text{pop}\  \mathrm{argmax}_{H_a}(fringe_a)$

    \State Sample $\mathbf{y}_{d,1},\ldots,\mathbf{y}_{d,k'}\ \sim p_{S_d}(\mathbf{y}\ |\ \mathbf{x}_{d,1}, \mathbf{x}_{d,2})$
    \State Sample $\mathbf{y}_{a,1},\ldots,\mathbf{y}_{a,k}\ \sim p_{S_a}(\mathbf{y}\ |\ \mathbf{c}_a, \mathbf{x}_a)$
    \For{each $\mathbf{y}_{d,j}$} 
    \If{$\mathbf{y}_d \notin seen_d\ \land V(\mathbf{x}_{d,1}, \mathbf{x}_{d,2}, \mathbf{y}_d)$}
        \State $seen_d \gets seen_d\ \cup\ \{\mathbf{y}_d\}$
        \State $fringe_d \gets fringe_d \cup \mathrm{pairs}(\mathbf{y}_d, seen_d)$
        \State $fringe_a \gets fringe_a \cup \mathrm{pairs}(\mathbf{y}_d, seen_a)$
    \EndIf
    \EndFor

    \For{each $\mathbf{y}_{a,j}$}
    \If{$\mathbf{y}_a \notin seen_a\ \land V(\mathbf{x}_a, \mathbf{y}_a, \mathbf{c}_a)$}
        \State $\textbf{yield}\ \mathbf{y}_a$
        \State $seen_a \gets seen_a\ \cup\ \{\mathbf{y}_a\}$
        \State $fringe_a \gets fringe_a \cup \mathrm{pairs}(\mathbf{y}_a, seen_d)$
    \EndIf
    \EndFor

\EndWhile
\end{algorithmic}
\label{alg:adgv}
\end{algorithm}

Our search relies on several modules, first selecting steps to take, then sampling generations from the different step types, validating generations, and finally populating the fringe with new generations.  The search algorithm is outlined in Algorithm \ref{alg:adgv}. The search operates over two fringes, an abductive and deductive fringe, which it will process in an interleaved fashion while adding new work items to both fringes. We allow the search to iterate until a specified number of steps $\mathrm{maxSteps}$ is reached. 


\paragraph{Prioritizing the Fringe: Learned heuristic models}
During search, we order the entries in the deductive fringe according to the Learned (Goal) heuristic model from \citet{bostrom2022natural}.  For the abductive fringe, however, we train a custom learned heuristic.\footnote{Note that adding goal conditioning to an abductive heuristic does not make sense as the model typically \emph{already} has knowledge of the goal in its inputs.}

To train the abductive heuristic, we produce a pool of positive abductive steps from the gold EntailmentBank train dataset by selecting an arbitrary intermediate step and pairing each of its inputs with the step's conclusion to yield a single positive example. We also produce negative samples by pairing an intermediate conclusion $\mathbf{c}$ and an arbitrary premise or other intermediate conclusion that is not part of $\mathbf{c}$'s subtree (previous inputs). The heuristic model is an instance of DeBERTa-v3 Large finetuned on all positive and negative samples. Further details are in the appendix.

\paragraph{Generating and Filtering} We allow for multiple generations to be sampled per step to fully explore the search space; however, this may lead to either invalid or redundant generations that need to be pruned.  A combination of validators $V(inputs, \mathbf{y}_i)$ remove any generations that do not meet a set of criteria, pruning their branch in the search space.  The fringe is then populated using the valid generations following the rules in Table~\ref{tab:allowed_steps}.

Our core validation methods to ensure logical correctness rely on a notion of \textbf{round-trip consistency}: we want deductive generations to work in reverse when plugged into the abductive model, and vice versa. More specifically, our \textbf{Deductive Agreement} module validates abductive steps, ensuring that the abductive generation (when combined with its input premise) produces the original conclusion.  For example, the abductive step $(\mathbf{c}, \mathbf{x}\ {\rightarrow}\ \mathbf{x'})$ is validated by taking the corresponding deductive step $(\mathbf{x}, \mathbf{x'}\ {\rightarrow}\ \mathbf{c}')$. The validator then checks that the scoring metric $E(\mathbf{c}, \mathbf{c'})$ is above a set threshold $t_{d}$.

The \textbf{Abductive Agreement} validator ensures that each input of a deductive step can be recovered using the output of the deductive step and the other input.  For example, the deductive step $(\mathbf{x}_1, \mathbf{x}_2\ {\rightarrow}\ \mathbf{c})$ is validated by taking two corresponding abductive steps $(\mathbf{x}_1, \mathbf{c}\ {\rightarrow}\ \mathbf{x}_2')$ and $(\mathbf{x}_2, \mathbf{c}\ {\rightarrow}\ \mathbf{x}_1')$.  The scoring metric is then checked for the two pairs $E(\mathbf{x}_1, \mathbf{x}_1')$ and $E(\mathbf{x}_2, \mathbf{x}_2')$. Both generated inputs' scores must be above a threshold $t_{a}$ for the output to be considered valid.

\paragraph{Other Validation Methods}
We also used two other validators: de-duplication and consanguinity thresholding. De-duplication removes any non-unique outputs as well as any output that is copied directly from the inputs of the step. Consanguinity thresholding looks at the ``ancestry'' of a generation up to depth $\eta$ and blocks generating from any pair that shares a given statement in their ancestry. We set $\eta=1$ to prevent combination of two of the same statement; higher thresholds did not help.

\subsection{Premise Recovery Scoring}
\label{sec:premise_recovery_scoring}

When the search concludes, we score each abductive generation $\mathbf{x'}$ to test for the recovery of $\mathbf{x}_m$ through a scoring metric $E(\mathbf{x'}, \mathbf{x}_m)$ which we then filter to candidates that pass a threshold $t_m$. To score each abduction, our system uses a harmonic mean $s = E(\mathbf{x}_m, \mathbf{x'}) = \frac{2 s_r s_e}{s_r + s_e}$ of $s_r = \textrm{ROUGE-1} (\mathbf{x}_m, \mathbf{x'})$ and an entailment scoring $s_e = entailed(\mathbf{x'}, \mathbf{x}_m)$ according to an entailment model. Every $\mathbf{x'}$ that recovers $\mathbf{x}_m$ has exactly one corresponding derivation that entails the goal, so we can associate it with a deductive proof tree. 

\subsection{Re-Ranking Proofs}
\label{sec:reranker}

Each proof found is re-ranked using the average deductive agreement score for every step in the proof using the validator.  The score is calculated on a single step $T_i = (\mathbf{x}_1, \mathbf{x}_2 \rightarrow \mathbf{c})$ by recreating $\mathbf{c}$ using the deductive step model $\mathbf{c'} = S_d(\mathbf{x}_1, \mathbf{x}_2) \rightarrow \mathbf{c'}$.  We then test $\mathbf{c'}$ for entailment of the original step's conclusion $s = entail(\mathbf{c'}, \mathbf{c}$) and taking the entailment probability as a score. Averaging these probabilities across all steps, $score = \frac{1}{n}\sum^n_{i=0}s_i$ where $n$ are the total number of steps in the proof, favors proofs with both deductive and abductive steps that verify deductively and minimizes the expected fraction of errors in the proof.

\section{Everyday Norms: Why Not?}

\begin{figure}[t!]
    \centering
    \includegraphics[width=1.0\columnwidth]{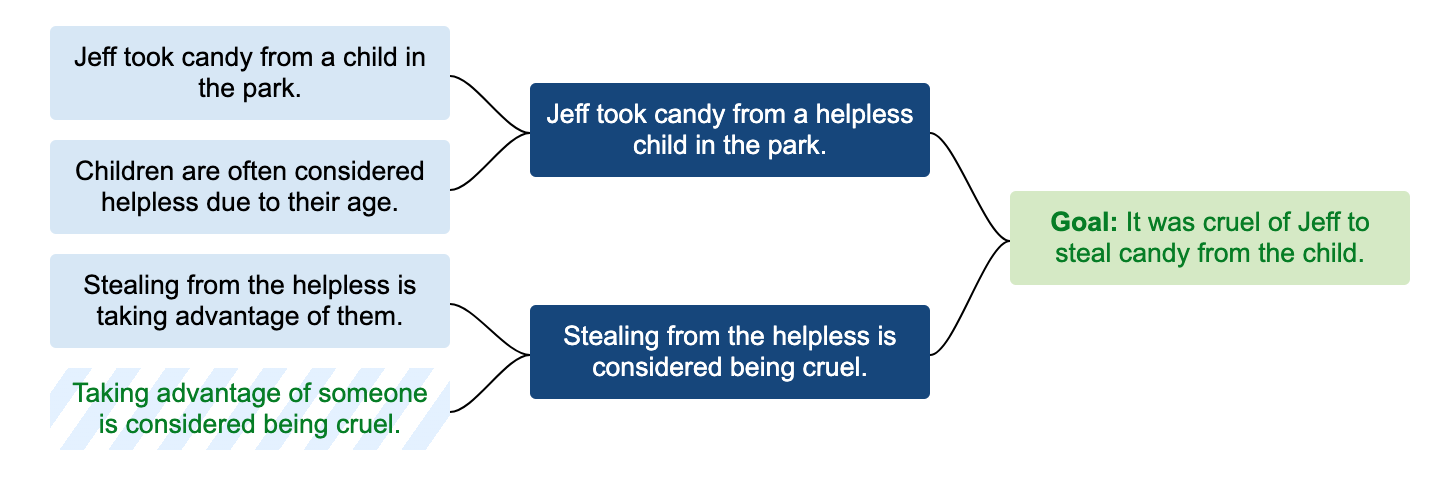}
    \caption{An entailment tree  example from the ENWN dataset.  Light blue blocks are premises (given as input to the step models), with the last being the gold missing premise, and dark blue blocks are gold intermediate steps (used during training but hidden during inference). The green bock is the goal statement.}
    \label{fig:enwn_example}
\end{figure}

To evaluate our method, we need data consisting of entailment trees $T$ as shown in Figure \ref{fig:enwn_example}. EntailmentBank \cite{dalvi-etal-2021-explaining} is the only existing dataset suitable for this evaluation; however, it is limited to the elementary science domain and we found that step models can often elide minor steps such as synonym replacements, making many instances easy to solve.

We collect a new English-language dataset called \emph{Everyday Norms: Why Not?} (ENWN) describing why an action is or isn't appropriate given a set of circumstances and a set of assumed norms.\footnote{We note that unlike Delphi \cite{delphi}, all of the assumptions here are made explicit in the inputs (\emph{taking your neighbor's property without asking permission is stealing}, \emph{stealing is wrong}) rather than relying on the system's priors. Our emphasis is on benchmarking the ability of systems to produce conclusions given stated premises, not trying to automate moral judgments per se \cite{talat-machine-ethics}.} ENWN consists of 100 entailment trees annotated by the first two authors of this paper. Each example considers a unique situation, providing an ethical judgement and its justification in the form of an entailment tree. Premise statements include both information about a situation as well as ethical norms. Intermediate steps are written to be similar in form to those in the EntailmentBank dataset, with the exception that all steps have two input statements.

Examples are given in Figure~\ref{fig:enwn_example} and Appendix \ref{sec:ethicstest_examples}. ENWN trees are slightly larger than EntailmentBank trees, with an average of 4.71 steps in comparison to EntailmentBank's 4.26. Anecdotally, we note that most steps in ENWN cannot be easily elided as they do not involve premises expressing trivial identities, such as ``\emph{Green is a color},'' which occur often in EntailmentBank.

\section{Experimental Setup}

We evaluate our models on the premise recovery task according to the two criteria stated previously, recall of missing premises and validity. Recall of missing premises, which we refer to as coverage, is defined using our recovery scoring $E(\mathbf{x}_m, \mathbf{x'}) \geq 0.7$. An example only has to produce one tree containing the missing premise to be counted towards the coverage metric. We use human evaluation to evaluate validity.

We evaluate on the English-language EntailmentBank \cite{dalvi-etal-2021-explaining} test set and our new \emph{Everyday Norms: Why Not?} (ENWN) dataset. To control for tree depth, our test examples are produced by slicing each full entailment tree into \emph{treelets} and removing a single premise from each treelet. Slicing trees allows us to create settings of varying difficulty (deeper treelets being more difficult) and since each treelet has at least two premises we can generate many individual examples. For our evaluations in Table \ref{tab:main_metrics} we use a random sampling of 100 treelets for both the EntailmentBank and ENWN datasets.

We compare various models, including End-to-End (E2E), Deductive only (DG), Abductive only (AG), Abductive and Deductive (ADG), and finally our full model, Abductive and Deductive with Validation (ADGV). We now proceed to describe these models.





\subsection{Baselines}
\label{sec:DG_description}

\paragraph{Deductive Generation Only (DG)}

The first baseline we compare against is the deduction system of \citet{bostrom2022natural}. We will simply use this system as originally specified, applying it to the incomplete premises to see if the missing premise can be inferred through deduction alone.

\begin{table*}[h!]
    \centering
    \small
    \begin{tabular}{r c c c c c c c c c c}
        \toprule
        & \multicolumn{5}{c}{\textbf{Entailment Bank}} & \multicolumn{5}{c}{\textbf{Everyday Norms: Why Not?}} \\
        \textbf{System} & \textbf{D1} & \textbf{D2} & \textbf{D3} & \textbf{D4} & \textbf{Full} & \textbf{D1} & \textbf{D2} & \textbf{D3} & \textbf{D4} & \textbf{Full} \\
        \midrule
        DG   & N/A  & 0\%  & 0\%  & 0\%  & 0\%  & N/A & 0\% & 0\%  & 0\% & 0\% \\
        AG   & \textbf{76\%} & 43\% & 27\% & 22\% & 37\% & \textbf{57\%} & \textbf{16\%} & \textbf{13\%} & 13\% & \textbf{12\%} \\
        ADG  & \textbf{76\%}  & \textbf{46\%} & \textbf{28\%} & \textbf{25\%} & \textbf{39\%} & \textbf{58\%} & \textbf{16\%}   & \textbf{11\%} & \textbf{14\%} & 13\% \\
        ADGV & 53\% & 20\% & 10\% & 8\% & 14\% & 30\% & 3\% & 4\%  & 2\% & 2\% \\\midrule
        E2E & 73\% & 56\% & 53\% & 46\% & 56\% & 41\% & 28\% & 19\% & 21\% & 21\% \\
        \bottomrule
    \end{tabular}
    \caption{
        The percentage of premises recovered across both datasets stratified by the depth of trees. Each D$k$ setting is restricted to trees of only that depth, with full containing full trees that represent all depths (but not a union of all other settings). The E2E baseline is separated out as it does not produce proofs along with its generations.
    }
    \label{tab:main_metrics}
\end{table*}


\paragraph{Abductive Generation Only (AG)}

This model only uses abductive generation. Though this can be effective for certain tree structures and small trees, it cannot generate any intermediate steps requiring forward inference as in Figure~\ref{fig:abduction_intro}.

\paragraph{End to End (E2E)}

Finally, we compare against an end-to-end model that generates a premise conditioned on a set of premises and a goal.  We use T5 3B fine-tuned on an adapted EntailmentBank dataset with appropriately constructed training examples; more details are in Appendix \ref{sec:appendix}. Note that this model \textbf{does not} generate a proof and \emph{only} infers the premise, which we will see can lead to reasoning shortcuts.

\subsection{Implementation Details}

We run our search for $\mathrm{maxSteps}$ timesteps. Each system is given the same number of backward steps to control for the steps that can actually generate the missing premise (2, 4, 8, 16, 25 for depths 1, 2, 3, 4, and all respectively). A forward step budget is added on top of this (2, 4, 8, 16, and 25 for depths 1, 2, 3, 4, and all respectively), which does increase wall-clock time for two-fringe models (ADG and ADGV) in relation to single fringe models (AG and DG). All models are allowed to sample multiple generations; for abductive steps we sample 40 generations and for deductive steps we sample 10 generations. 

Runtimes lengthen as the total number of steps increase and the total number of generations sampled increase.  For the largest model (ADGV) with 50 total steps, 40 abductive generation samples per step and 10 deductive generation samples per step, examples are completed in 1 to 2 minutes on average.

Our sequence-to-sequence models are instances of T5 3B \cite{raffel-etal-2020-t5}. Our entailment models and learned heuristic models use DeBERTa Large \cite{he2021deberta} with 350M parameters. All models are implemented with the Hugging Face \texttt{transformers} library \cite{wolf-etal-2020-transformers}. Further details including fine-tuning hyperparameters are included in the appendix.


\paragraph{Premise Recovery Scores}

All of the $entailed(\mathbf{x}, \mathbf{y})$ calls performed during the search use the same EBEntail-Active model as in \citet{bostrom2022natural}.  We define the \emph{rightward entailment} score in our work as $score = entailed(\mathbf{x}_m', \mathbf{x}_m)$. This entailment can be read as the generated missing premise entailing the actual missing premise.  We empirically found this to agree best with our annotations, as discussed in Section \ref{sec:coverage}, and used rightward for our results.

\begin{table*}[h!]
    \centering
    \small
    \begin{tabular}{r c c c c c c c c c c}
        \toprule
        & \multicolumn{4}{c}{\textbf{Entailment Bank}} & \multicolumn{4}{c}{\textbf{Everyday Norms: Why Not?}} \\
        \textbf{System} & \textbf{Count} & \textbf{Len} & \textbf{Score} & \textbf{P Recall} & \textbf{Count} & \textbf{Len} & \textbf{Score} & \textbf{P Recall} \\
        \midrule
        AG   & 10.30 & 3.99 & 43\% & 59\% & 6.67 & 3,.51 & 24\% & 32\%\\
        ADG  & 9.97  & 7.74 & 50\% & 66\% & 3.46 & 6.38 & 29\% & 48\%\\
        ADGV & 7.43  & 4.11 & 80\% & 82\% & 2.50 & 8.12 & 83\% & 79\%\\
        \bottomrule
    \end{tabular}
    \caption{We compare ADGV with two baselines, ADG and AG, on 4 metrics across both the Entailment Bank and ENWN datasets.  \emph{Count} is the number of proofs solved on average per tree.  \emph{Len} is the length of the proofs on average (how many deductive and abductive steps).  \emph{Score} is the average of the entailment probabilities from the deductive agreement score, see section \ref{sec:reranker}.  Finally, \emph{P Recall} is premise recall: the percentage of the original premises used in the proof. On average the ADGV algorithm produces fewer proofs, but higher scoring proofs that use more of the original premises in its proofs than the baseline methods.}
    \label{tab:tree_metrics}
\end{table*}

\section{Results}

\begin{table}[ht]
\renewcommand{\tabcolsep}{1.0mm}
    \centering
    \small
    \begin{tabular}{c c}
        \toprule
        \textbf{Model} & \textbf{Recovered} \\
        \midrule
        E2E & 56\% \\
        E2E w/o Goal & 32\% \\
        \bottomrule
    \end{tabular}
    \caption{We compare the End-to-End (E2E) model with a variant of the E2E model where no goal is given.  Logically, without the goal, it should be impossible to derive the correct missing premise as the space of all generations is too large.  Despite this large space, the E2E w/o Goal model is capable recovering the premise 32\% of the time, illustrating the existence of shortcuts the model can exploit.}
    \label{tab:premise_algebra}
\end{table}

As our chief goal is to infer missing premises, we begin with premise recovery (coverage) results, shown for all baselines and our best models across both datasets in Table~\ref{tab:main_metrics}. We then discuss human evaluation of both step validity (Table~\ref{tab:stepv_tab}) and coverage (Table~\ref{tab:coverage_tab}).

\subsection{Coverage Results}

\paragraph{Abductive generations are required for recovering premises.}

DG cannot recover any of the premises at any level of depth,\footnote{Because the Forward step model requires at least 2 premise statements to perform a step, the model was not run in the D1 setting because those trees only have 1 premise.} illustrating that these premises truly are unstated assumptions not derivable through forward inference. 

\paragraph{Using deductive steps generally improves coverage (and validity).}

AG is capable of producing the missing premise nearly as often (and sometimes more so) than ADG.  However, because the re-ranking algorithm in Section~\ref{sec:reranker} favors steps with high deductive agreement, ADG produces slightly higher quality proofs in general, shown in Table \ref{tab:tree_metrics}'s \emph{Score} column.



\paragraph{Validators vastly improve quality at the cost of recall}  

Although using validators produces far fewer proofs in Table~\ref{tab:main_metrics}, the quality of proof trees is vastly improved in the ADGV setting. We study the statistics of these generated trees in Table \ref{tab:tree_metrics}.  Because there are not actually many valid ways to recover a missing premise, lower proof counts typically indicate more reliable proofs. Shorter proofs also tend to be more consistent with those in the gold entailment trees.  Score is the deductive agreement score used to rank the proofs, with higher scores indicating better validity. Finally, Premise Recall (P Recall) is the percentage of the original premises used in the proof.  High Premise Recall indicates that more of the input was used to derive the missing statement which indirectly leads to better quality and indicates less hallucination.

Appendix~\ref{sec:example_proofs} shows examples of successful and unsuccessful proofs from this method. These illustrate the difficulty of our dataset instances, highlighting how we need to not only chain together the correct inferences and produce the correct statement but also do so within the search budget. Exhaustive search over the space of natural language statements leads to an exponentially large fringe; however, overly heavy filtering may remove a precisely-worded intermediate conclusion needed to recover the missing premise exactly. Finding a balance is a key challenge with stronger methods. 

\paragraph{While E2E can recover many premises, it does not construct proofs and uses shortcuts}

In nearly every depth setting, the E2E model recovers a higher number of premises than our methods.  However, the mechanisms that produce these generations can be unsound. 
Often, when abduction is performed, the level of specificity to abstract or retain is underspecified (as mentioned in Section~\ref{sec:methods_intro}). The E2E model is able to learn these levels of specificity and perform a ``premise algebra'' from priors in the training data that the step generation baselines cannot exploit (see examples in Appendix \ref{sec:E2E_no_goal_examples}). That is, the model can identify keywords that are systematically missing from examples and infer that the missing premise must use them. 

Table \ref{tab:premise_algebra} shows an experiment in which the E2E model is given a set of incomplete premises without the goal and is asked to produce the missing premises.  We find that this E2E without Goal model is capable of solving 32\% of the examples showing that more than half of the examples solved by the E2E model in Table \ref{tab:main_metrics} could have been solved using premise algebra shortcuts. In contrast, our model cannot exploit these shortcuts. 


\paragraph{ENWN is a challenging dataset for future work} Even the ``premise hacking'' E2E model only achieves around 20\% recovery of missing premises on the full setting. Producing a valid tree that recovers the correct premise is out of range of our current models given our computation budget. We expect scaling the sizes of our models and using improved filtering during search to prioritize the right branches may lead to improvements.

\subsection{Human Step Validity Evaluation}
\label{sec:step_validity}

Beyond coverage, we want to ensure that our models are taking sound abduction steps, which can also help evaluate whether the model is able to make valid inferences even if the missing premise is not recovered.

We collected steps in two settings: steps from the top ranking proof in cases where the missing premise was recovered (\emph{Top Proof}) and steps in the search state explored at any time from any example (\emph{All}). We then labeled these steps for validity. Soundness is defined as whether the abductive inference yields (1) a true new premise (2) that validates in the forward deductive inference.

\begin{table}[t!]
\renewcommand{\tabcolsep}{1.0mm}
    \centering
    \small
    \begin{tabular}{@{}r |c c c |c c c c@{}}
        \toprule
        & \multicolumn{3}{c|}{\textbf{Top Proof}} & \multicolumn{3}{c}{\textbf{All}} \\
        \textbf{Model} & \textbf{Valid} & \textbf{VND} &  \textbf{Invalid} & \textbf{Valid} & \textbf{VND} & \textbf{Invalid} & \\
        \midrule
        ADG & 52.8\% & 11.1\% & 36.1\% & 40.4\% & \phantom{0}6.4\% & 53.2\%\\
        ADGV & \textbf{87.0\%} & \textbf{\phantom{0}4.3\%} & \textbf{\phantom{0}8.7\%} & \textbf{72.5\%} & \textbf{\phantom{0}0\%} & \textbf{27.5\%}\\
        \bottomrule
    \end{tabular}
    \caption{Results of manually annotating a total of 200 reasoning steps for validity sampled across two models, ADG and ADGV, and two settings, Top Proof and All, (50 samples per pair) showing that ADGV yields significantly higher quality.}
    \label{tab:stepv_tab}
\end{table}

The label set includes Y (valid), N (invalid), VND (``valid but not deductive'': a true premise that doesn't result in a valid forward deduction).  Only examples labeled Y are considered a \emph{valid} step. Ties between valid and invalid annotations favors invalid. Agreement across the multiple labels (Cohen’s $\kappa$) was $0.48$.

As shown in Table~\ref{tab:stepv_tab}, on average, using validators produces nearly twice as many valid steps while searching for a proof.  Because the proofs are re-ranked once found, the gap between ADG and ADGV in the Top Proof setting is not as dramatic, but still shows a major improvement in creating sound proofs. Having valid steps in complete proofs is important for soundness, but having more valid steps anywhere in the search state demonstrates that the ADGV search explores valid branches of reasoning more often than not.

\subsection{Human Coverage Evaluation}
\label{sec:coverage}

Our coverage numbers in Table~\ref{tab:main_metrics} are an automatic estimate. We undertook additional human validation to ensure that these numbers are representative of actual premise recovery rates.

We sampled 100 steps that were identified as having recovered a premise. Three of the authors then annotated each step as truly recovering the missing premise based on either exhibiting mutual entailment ($\mathbf{x}_m' \leftrightarrow \mathbf{x}_m$) or more specific premises ($\mathbf{x}_m' \rightarrow \mathbf{x}_m$), see Figure \ref{fig:under_and_entail} for an example.  Statements that were more general but did not entail the missing premise were relatively rare and were not considered correct (although they can be valid abductive inferences in some cases), along with other unrelated or bad cases.  Ties between the annotators favored the negative (the premise was not recovered); however, annotator agreement was reasonably high with Cohen’s $\kappa$ at $0.74$.

Table \ref{tab:coverage_tab} measures the premise recovery agreement  (\emph{coverage}) of the ADG and ADGV systems with manual annotators. We note that the majority of premises marked as recovered by the system are valid missing premises, supporting the validity of our results in Table~\ref{tab:main_metrics}. However, we see that the validated results in ADGV tend to align better with human judgments by $14\%$; this casts the recovery results of Table~\ref{tab:main_metrics} in a more favorable light for the ADGV system.

\begin{table}[t!]
\renewcommand{\tabcolsep}{1.0mm}
    \centering
    \small
    \begin{tabular}{c c}
        \toprule
        \textbf{Model} & \textbf{Recovered} \\
        \midrule
        ADG & 68\% \\
        ADGV & \textbf{82\%} \\
        \bottomrule
    \end{tabular}
    \caption{Fraction of automatically-identified recovered premises that our human labeling identified as correct from two of our systems.}
    \label{tab:coverage_tab}
\end{table}

\subsection{Error Analysis: ADGV}
\label{sec:err_step}

\paragraph{Underspecification}

Although validation can help avoid abductive underspecification, the validation models can fail to filter invalid steps. For example, $\mathbf{x}_a\ =$ ``\emph{A reptile does not have fur.}'' and $\mathbf{g}\ =$ ``\emph{Animals in the taxonomic groups bird and reptile do not have fur.}'' combine together produce the abductive generation ``\emph{Birds do not have fur.}'' Although this inference passes validation, if we attempt to recreate the goal through forward deduction, we would fail as information about taxonomic groups of animals is not specified. The validator thresholds could be changed to filter this, but this is a challenging case anyway as it is not obvious how to phrase an abductive generation to yield the correct result here.

\paragraph{Cascading errors}
There is no way for ADGV to test for fallacious generations or false premises. For example, if $\mathbf{x}_a\ =$ ``\emph{A plant is a kind of living thing.}'' and $\mathbf{c}\ =$ ``\emph{Grass and a cow are both a kind of plant.}'', $\mathbf{c}$ is a false statement, but the abductive step model can still produce a valid generation ``\emph{Grass and a cow are both living things.}''. However, any proof generated that includes this step would be unsound because $\mathbf{c}$ is false. 

\paragraph{Premises that subsume their conclusions}

If a premise statement $\mathbf{x}_a$ includes a conclusion $\mathbf{c}$, there is nothing to infer from the resulting abductive step that would be meaningful.  However, the abductive heuristic can still select these steps and generate abductive inferences that bypass validation. For example, if $\mathbf{x_a}\ =$ ``\emph{A substance is highly reflective, able to conduct electricity, and have high melting points.}'' and $\mathbf{c}\ =$ ``\emph{The substance has high melting point.}'', $\mathbf{x_a}$ entails $c$ on its own (as well as additional information), leading the step model to generate $\mathbf{x_b}\ =$ ``\emph{The substance is highly reflective and able to conduct electricity.}''  Although $\mathbf{x_b}$ may be true, it is not strictly an abduction, and as an independent statement will tend to pollute the search on the next step. Preventing premises from combining with conclusions they already entail could reduce search state complexity and increase step validity, but this is left for future work.

\section{Related work}

Our work stems from well established models in the question answering domain \cite{rajpurkar-etal-2016-squad, yang-etal-2018-hotpotqa, kwiatkowski-etal-2019-natural}.  Specifically, models have often looked at either generating the correct answer or selecting statements from a set to derive an answer in a ``multi-hop'' manner \cite{chen2021multihop, min-etal-2019-multi, nishida-etal-2019-answering}.  Although discriminative models select evidence for their answers, there is little reasoning being exposed making it hard to detect affordances taken by the end-to-end approaches \cite{hase-bansal-2020-evaluating,BansalEtAl2021}.

Recently, step-by-step models have been used to create entailment trees that expose a model's reasoning down to individual deductive operations \cite{bostrom2022natural, dalvi-etal-2021-explaining, ribeiroIRGR}. Some with the ability to perform backward inferences have also been introduced \cite{hong2022metgen, qu2022interpretable}. However, these methods focus on entailing a goal rather than recovering missing evidence. Other work has explored validating step model generations \cite{Yang2022GeneratingNL}, but to our knowledge none have used abductive and deductive step models to mutually validate each other.

Chain-of-thought prompting techniques have been used to conduct step-by-step reasoning by eliciting intermediate steps from large language models \citep{cot, selection_inference}, but these have been applied to other problems and some preliminary experiments indicate that they do not immediately work for our setting. A related method has been proposed which decomposes statements into inferred premises via backward inference \citep{maieutic}, although this approach does not simultaneously connect forward inferences from provided premises as our proposed method does.


\section{Conclusion}

In this work, we tackle the generation of missing premise statements in textual reasoning through the use of abduction. We introduce a new system capable of abductive and deductive step generation, which yields inferred missing premises while building a proof showing its reasoning.  Furthermore, we propose a novel validation method that reduces hallucination and other common failure modes in end-to-end and stepwise searches. Future work can improve our system by scaling up the models used, plus using additional notions of validation as discussed in the error analysis. We believe our overall framework can be a promising foundation for future reasoning systems.

\section{Limitations}

End-to-end models are able to produce a single generation per example reducing the time complexity for sufficiently small sets of premises.  Step-by-step models like our search procedure in this work are capable of handling sets of any size of premises for the search, but do increase the execution time per example, especially when using validators that require doing generation themselves. Nevertheless, validators do reduce the total time required for running a set of examples due to their ability of pruning the search space and thus removing numerous heuristic and generation calls.  With better heuristics and validators it may be possible to reduce the time complexity further, but that is left for future work.

Both the EntailmentBank and ENWN dataset were written in English and capture relatively limited domains of textual reasoning.  Different languages might introduce easier lexical patterns for abstraction though and could be a promising path forward. We believe ADGV and its variants should work on non-English languages, but testing this was left to future work.

ENWN draws on everyday ethical scenarios because this was a domain we found fruitful to exhibit the kind of reasoning our system can do. However, we do not follow in the steps of Delphi \cite{delphi} in making \emph{any} claims about its ability to make systems ethical or say anything about ``values'' encoded in pre-trained models. We do not support its use as part of any user-facing system at this time.

\section*{Acknowledgments}

This work was supported by NSF CAREER Award IIS-2145280, the NSF Institute for Foundations of Machine Learning,
ARL award W911NF-21-1-0009, 
a grant from Open Philanthropy, and gifts from Salesforce and Adobe. This material is also based on research that is in part supported by the Air Force Research Laboratory (AFRL), DARPA, for the KAIROS program under agreement number FA8750-19-2-1003. The views and conclusions contained herein are those of the authors and do not represent the views of DARPA or the U.S.~Government. Thanks to the anonymous reviewers for their helpful feedback.

\bibliography{custom}
\bibliographystyle{acl_natbib}
\newpage
\clearpage
\appendix

\begin{table}[t]
\renewcommand{\tabcolsep}{1.0mm}
    \centering
    \small
    \begin{tabular}{c c}
        \toprule
        \textbf{Step Signature} & \textbf{Step Type} \\
        \midrule
        $\mathbf{x}, \mathbf{x} \rightarrow \mathbf{y}_d$ & Deductive \\
        $\mathbf{x}, \mathbf{y}_d \rightarrow \mathbf{y}_d$ & Deductive \\
        $\mathbf{g}, \mathbf{x} \rightarrow \mathbf{y}_a$ & Abductive \\
        $\mathbf{g}, \mathbf{y}_d \rightarrow \mathbf{y}_a$ & Abductive \\
        $\mathbf{y}_a, \mathbf{x} \rightarrow \mathbf{y}_a$ & Abductive \\
        $\mathbf{y}_a, \mathbf{y}_d \rightarrow \mathbf{y}_a$ & Abductive \\
        \bottomrule
    \end{tabular}
    \caption{A list of possible input statement types each step model can take. $\mathbf{x}$ refers to a premise, $\mathbf{y}_d$ refers to an intermediate deductive conclusion, $\mathbf{g}$ refers to the goal, and $\mathbf{y}_a$ refers to an abductive hypothesis. Note that the deductive model can accept inputs in any order but the abductive model cannot, as the abduction operation is not commutative.  Also note that deductive outputs can be used as inputs to abductive steps, but not the other way around; allowing deductive steps to accept abductive generations could result in vacuous proofs. }
    \label{tab:allowed_steps}
\end{table}

\section{Implementation Details}
\label{sec:appendix}

All experiments were conducted using Hugging Face \texttt{transformers} version 4.20.0.

For all experiments in this paper a set of 3 Quadro 8000 GPUs with 48GB of RAM were used.

Model weights from \citet{bostrom2022natural} were used for the Deductive step model, Learned (Goal)+PPM heuristic model and the entailment model.

Default hyperparameters from HuggingFace are used if not otherwise specified for all Step models and the End-to-End model. No hyperparameters sweeps were conducted on these:

\begin{table}[h!]
\begin{tabular}{r c}
    \toprule
    \textbf{Hyperparameter} & \textbf{Value} \\
    \midrule
    Base model & \href{https://huggingface.co/t5-3b}{T5 3B} \\
    Total batch size & 8 \\
    Initial LR & 5e-5 \\
    Epoch count & 3 (early stopping on val. loss) \\
    \bottomrule
\end{tabular}
\caption{Abductive Step Model \texttt{transformers} default if unspecified)}
\end{table}

\begin{table}[h!]
\begin{tabular}{r c}
    \toprule
    \textbf{Hyperparameter} & \textbf{Value} \\
    \midrule
    Base model & \href{https://huggingface.co/microsoft/deberta-v3-large}{DeBERTa-v3 Large} \\
    Total batch size & 32 \\
    Initial LR & 2e-5 \\
    Epoch count & \begin{tabular}{@{}c@{}}2 \end{tabular} (early stopping on val. loss) \\
    \bottomrule
\end{tabular}
\caption{Abductive learned heuristic model fine-tuning}
\end{table}

\clearpage

\section{Everyday Norms: Why Not? Examples}
\label{sec:ethicstest_examples}

\begin{figure*}[b]
    \centering
    \vspace*{-2cm}
    
    \includegraphics[width=1.0\linewidth]{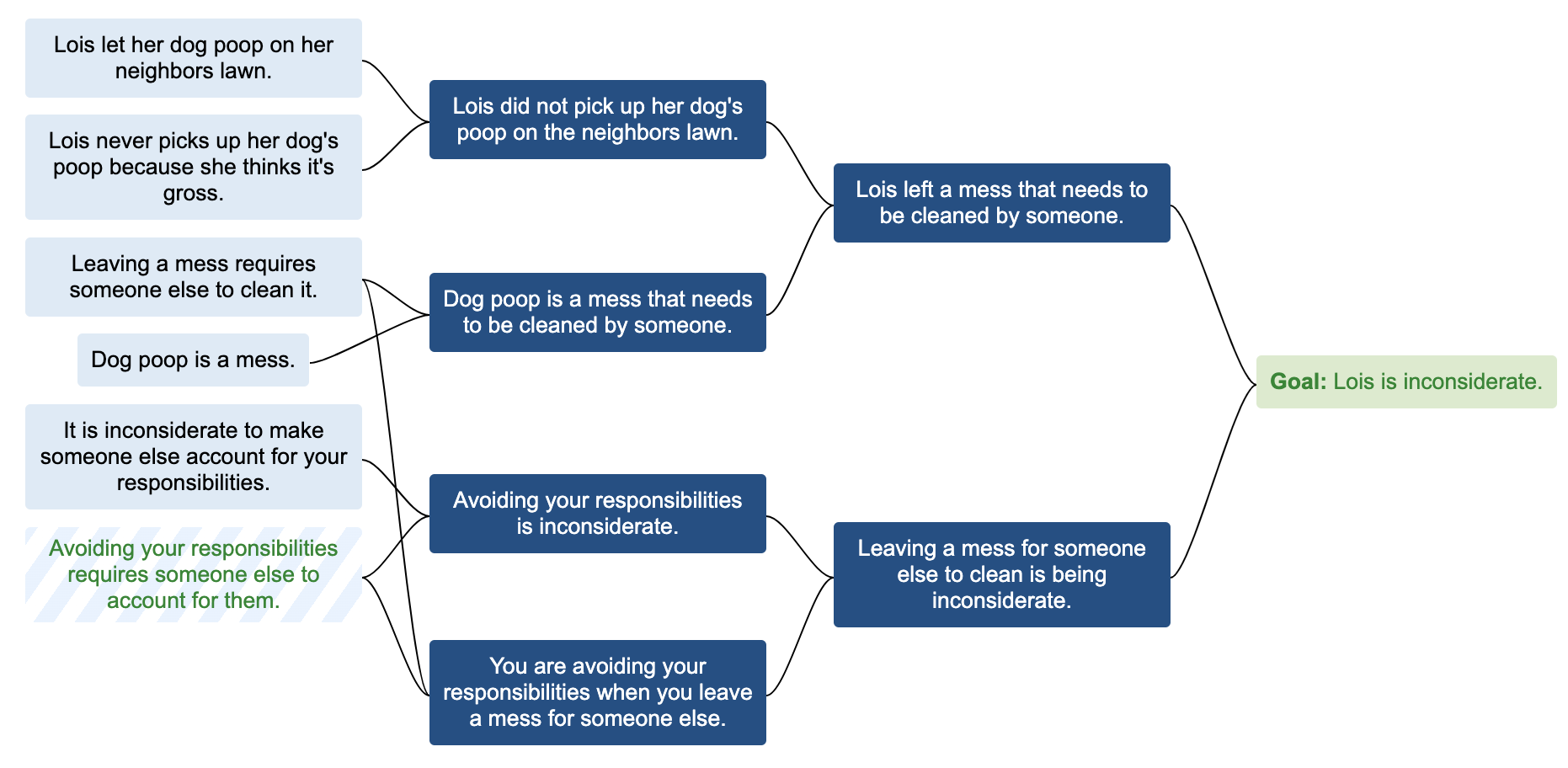}
    \includegraphics[width=1.0\linewidth]{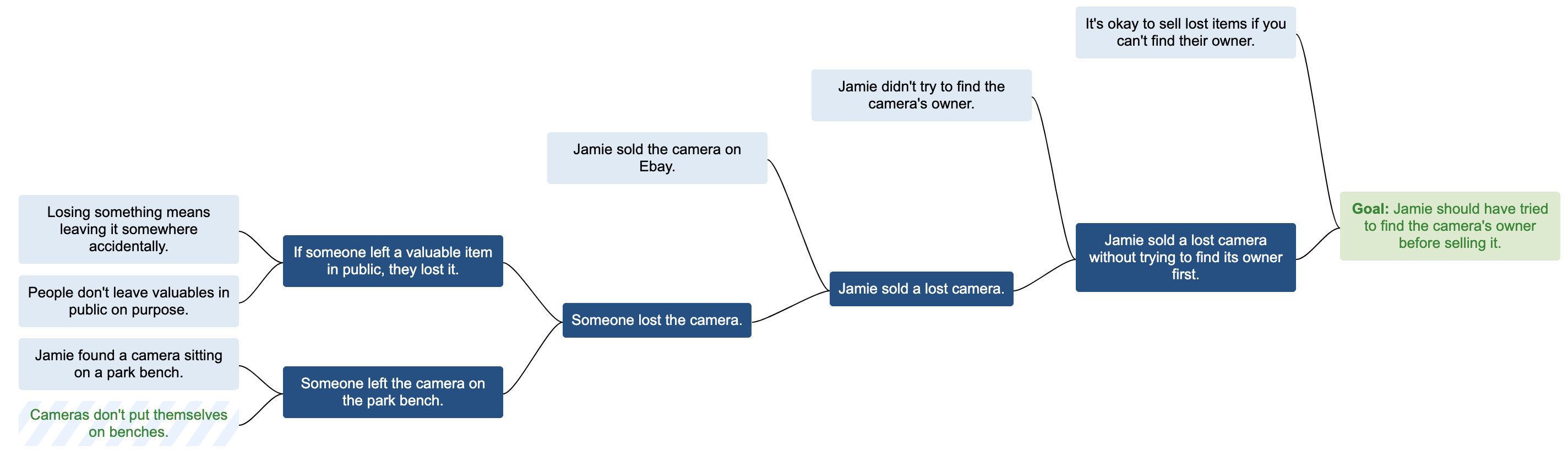}
    \includegraphics[width=1.0\linewidth]{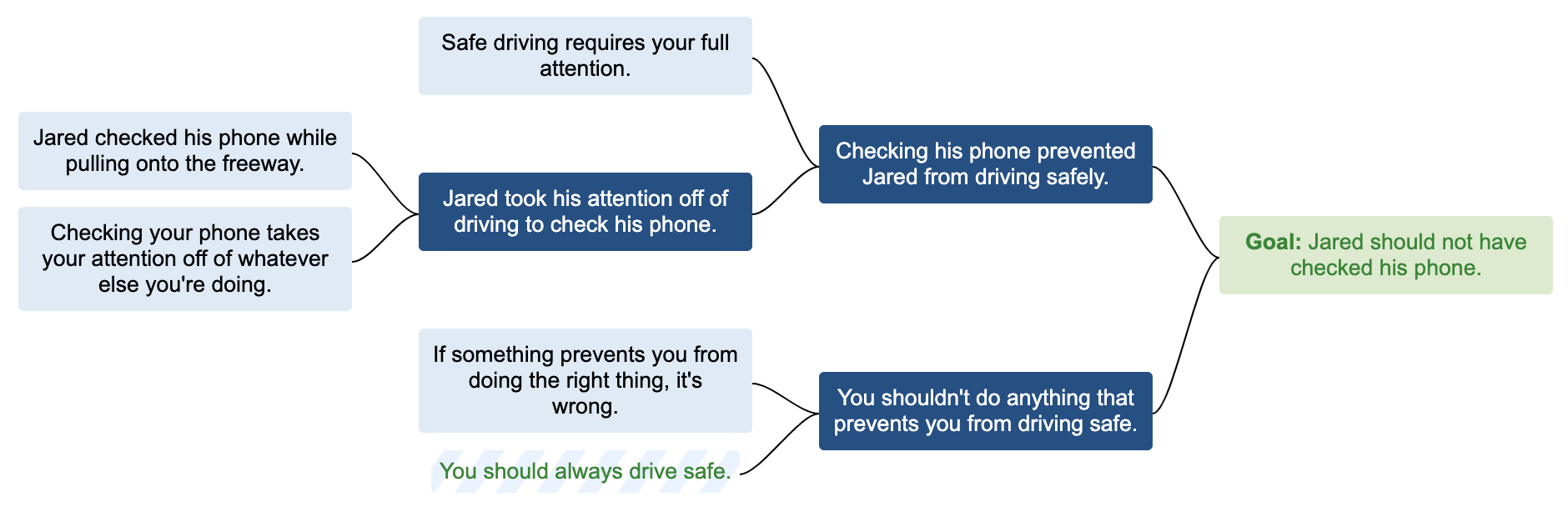}        
    \caption{ Three example entailment trees from the \emph{Everyday Norms: Why Not?} dataset.  Light blue boxes with white text are given premises, dark blue boxes are intermediate deductive steps, green boxes are the goal statements of the examples and striped blue boxes with green text are gold missing premises.  The arity for any intemerdiate step in \emph{Everyday Norms: Why Not?} is always two.}
    \label{fig:example_1_ENWN}
\end{figure*}

See Figure~\ref{fig:example_1_ENWN}.

\clearpage

\section{Premise Recovery Generation Examples}

See Table~\ref{tab:examples_of_premise_recovery}.

\begin{table*}[t]
    \centering
    \small
    \begin{tabular}{c | m{8em} | m{8em} m{2em} | m{8em} m{2em} | m{8em} m{2em}}
        \toprule
        \textbf{Depth} & \centering{\textbf{Gold}} & \multicolumn{2}{c|}{\textbf{AG}} & \multicolumn{2}{c|}{\textbf{ADG}}  & \multicolumn{2}{c|}{\textbf{ADGV}} \\

\midrule
1 & Some birds eat nectar. & some birds are animals that eat nectar. & 0.79 & birds can eat nectars. & 0.81 & birds that eat nectar eat nectar. & 0.72 \\
\midrule
1 & As the number of pathways increases, the traffic congestion in that area usually decreases. & as the number of pathways in an area increase, the traffic congestion in that area usually decreases. & 0.93 & as the number of pathways in an area increases, the traffic congestion in that area usually decreases. & 0.93 & as the number of pathways increase, the traffic congestion in that area usually decreases. & 0.99 \\
\midrule
1 & If fossils of an aquatic animal or plant are found in a place then that place used to be covered by water in the past. & so if fossils of aquatic animals is found in a place then that place used to be covered by water in the past. & 0.86 & . if fossils of aquatic animals are found in a place then that place used to be covered by water in the past. & 0.88 & if fossils of aquatic animals are found in a place then that place used to cover by water in the past. & 0.87 \\
\midrule
2 & Losing electrons causes the electrical charge of an object to be unbalanced. & as electrons go out of an object, the electrical charge in the object becomes unbalanced. & 0.71 & when an object loses electrons, the electrical charge of the object becomes unbalanced. & 0.81 & when objects lose electrons, the electrical charge of that object changes from balanced to unbalanced. & 0.76 \\
\midrule
2 & Acid rain causes water pollution. & acid rain is a source of water pollution. & 0.72 & acid rain lowers water quality. & 0.73 & acid rain is a pollutant. & 0.70 \\
\midrule
2 & Cold fronts cause thunderstorms as they pass by. & a cold front causes storms as it passes by. & 0.80 & cold fronts cause precipitation as they pass by. & 0.83 & cold fronts cause storms as they pass by. & 0.89 \\
\midrule
3 & Water absorbs solar energy in the water cycle. & water absorbs solar energy. & 0.74 & water absorbs solar energy. & 0.74 & water absorbs solar energy. & 0.74 \\
\midrule
3 & A fox is a kind of animal. & a fox is a kind of animal species. & 0.80 & fox is a kind of animal. & 0.95 & fox is a kind of animal. & 0.95 \\
\midrule
3 & Plants perform photosynthesis. & plants perform photosynthesis. & 0.99 & plants perform photosynthesis. & 0.99 & plants perform photosynthesis. & 0.99 \\
\midrule
4 & Cell division produces cells. & cell division produces cells. & 0.99 & cell division generates cells. & 0.84 & cell division produces cells. & 0.99 \\
\midrule
4 & Large birds are a kind of organism. & a large bird is an organism. & 0.73 & large birds are a kind of organism. & 0.99 & large birds are a kind of organism. & 0.99 \\

\bottomrule
    \end{tabular}
    \caption{A random sample of 11 abductive steps on the varying depth experiments from Table \ref{tab:main_metrics}.  The depth column corresponds with the depth on Table \ref{tab:main_metrics}. The gold column shows the original missing premise, then each following column represents one of the models showing it's best generation for that missing premise along with the $E(\mathbf{x'}, \mathbf{x})$ score.}
    \label{tab:examples_of_premise_recovery}
\end{table*}

 \section{E2E no goal premise recovery}
\label{sec:E2E_no_goal_examples}
In Table \ref{tab:e2e_no_goal_premise_recovery} we show three examples of the ablated model E2E w/o goal correctly generating the missing premise despite being given insufficient information to do so logically. This behavior is problematic as correctly identifying which premise to generate is a vast search space without the goal to direct the model --- clearly indicating that the E2E model has learned shortcuts in the data set and is taking advantage of them. Ideally, models would make sound inferences without using spurious patterns from the training dataset to create generations, which is exactly what our step models are designed to do. 

\begin{table*}[t!]
    \centering
    \small
    \begin{tabular}[h!]{m{12em} | m{8em} | m{8em} | m{18em}}
        \toprule
        \textbf{Hidden Goal} & \textbf{Missing Premise} & \textbf{E2E Output} & \textbf{Input} \\
        \midrule 
        The difference between the earth and the moon is that the moon revolves around a planet. & The moon orbits the earth. & The moon orbits the earth. & The sun is a kind of star. Revolving around means orbiting. The earth revolves around the sun. Earth is a kind of planet.  \\
        \midrule
        Earth is a celestial object located in the milky way galaxy. & Earth is a kind of planet. & Earth is a kind of planet. & A planet is a kind of celestial object / celestial body. Earth is located in the milky way galaxy. \\
        \midrule
        Dogs will inherit the color of fur from their parents. & Inheriting is when an inherited characteristic is copied / is passed from parent to offspring by genetics / dna. &  Inheriting is when an inherited characteristic is passed from parent to offspring by genetics / dna. & A dog is a kind of animal. Fur is often part of an animal. The color of / coloration of fur is an inherited characteristic. \\

        \bottomrule
    \end{tabular}
    \caption{Three examples of an End-to-End (E2E) model given only a subset of premises (no goal) generating a missing premise. In two out of the three example above the E2E model is capable of computing the missing premise word for word.}
    \label{tab:e2e_no_goal_premise_recovery}
\end{table*}

\clearpage 
\section{Example Proofs and Failed Searches}
\label{sec:example_proofs}

\begin{figure*}[b]
    \centering
    \vspace*{-2cm}
    
    \includegraphics[width=1.0\linewidth]{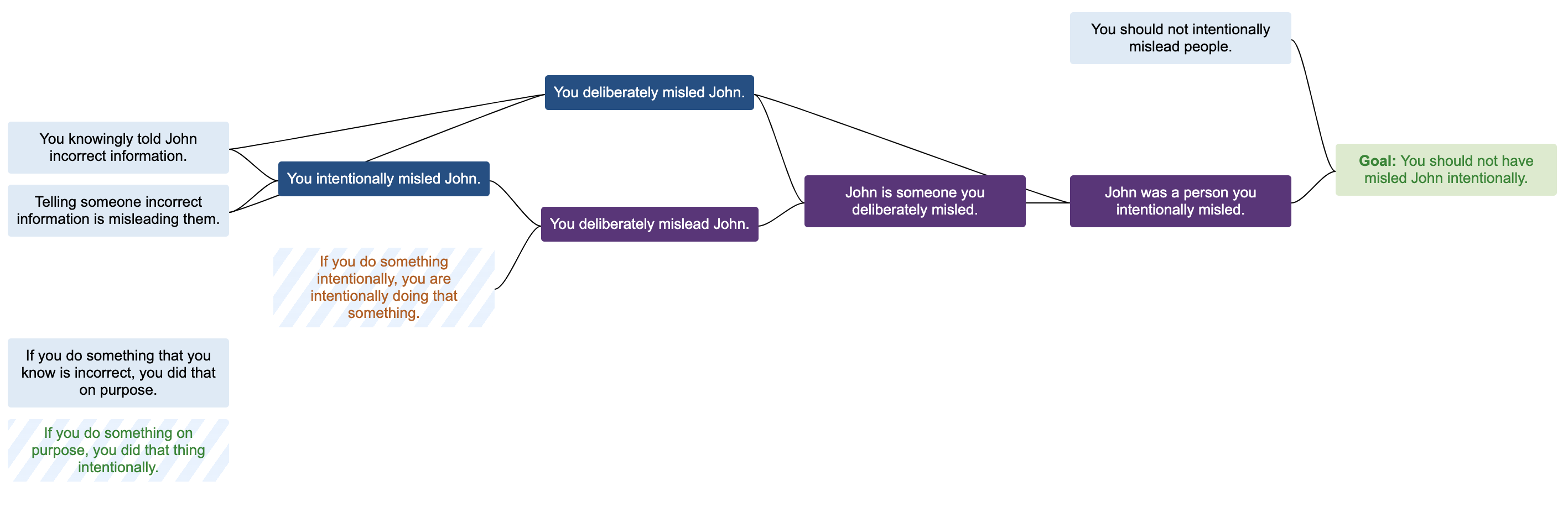}
    \includegraphics[width=1.0\linewidth]{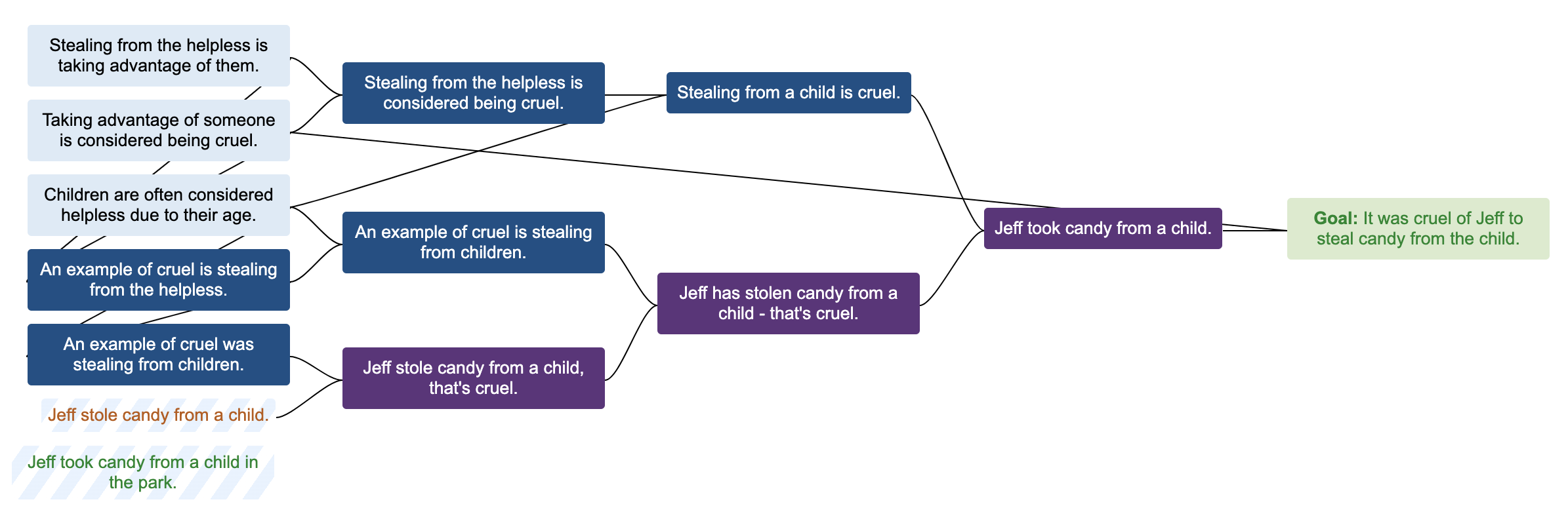}
    \includegraphics[width=1.0\linewidth]{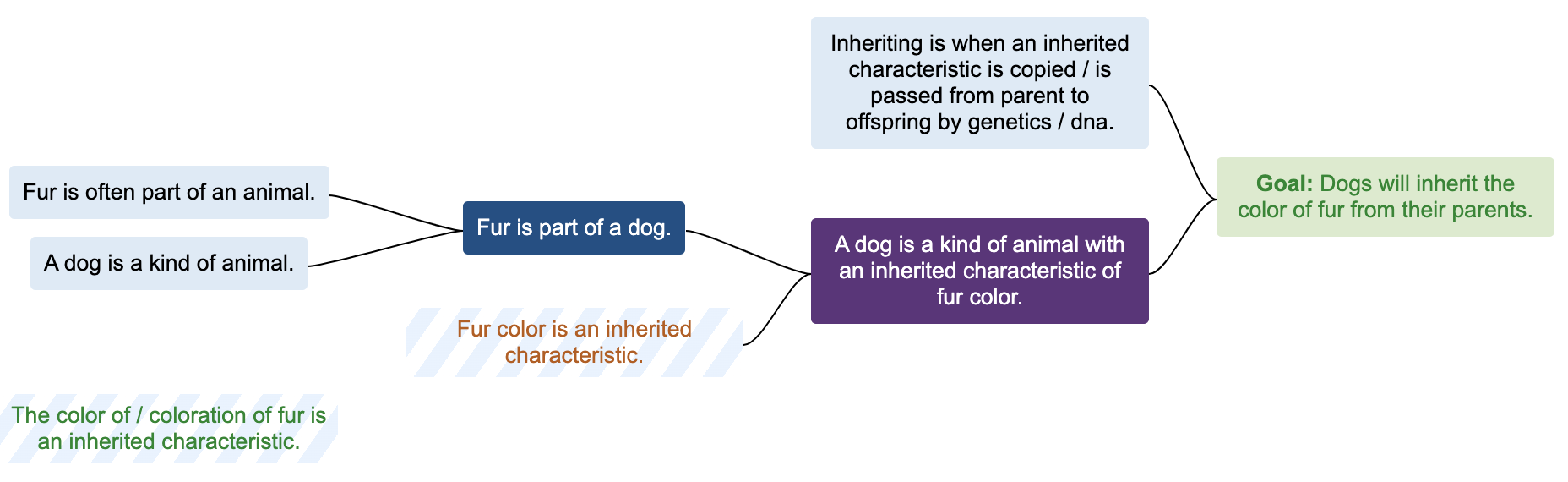}
    \caption{Example \textbf{successful proofs} using ADG from the Depth = all experiment. Boxes with blue stripes and orange text $\mathbf{x'}$ are generated premises from the abductive model where blue stripes with green text are the gold missing premise $\mathbf{x}_m$.  Light blue boxes are premises, dark blue are intermediate, purple are abductions, and green is the goal of the entailment tree. Note that the gold missing premise is never incorporated in the proof because we are trying to regenerate it through our step models.}
    \label{fig:adgv_example_proofs}
\end{figure*}

\begin{figure*}[b]
    \centering
    \vspace*{-1cm}
    
    \includegraphics[width=1.0\linewidth]{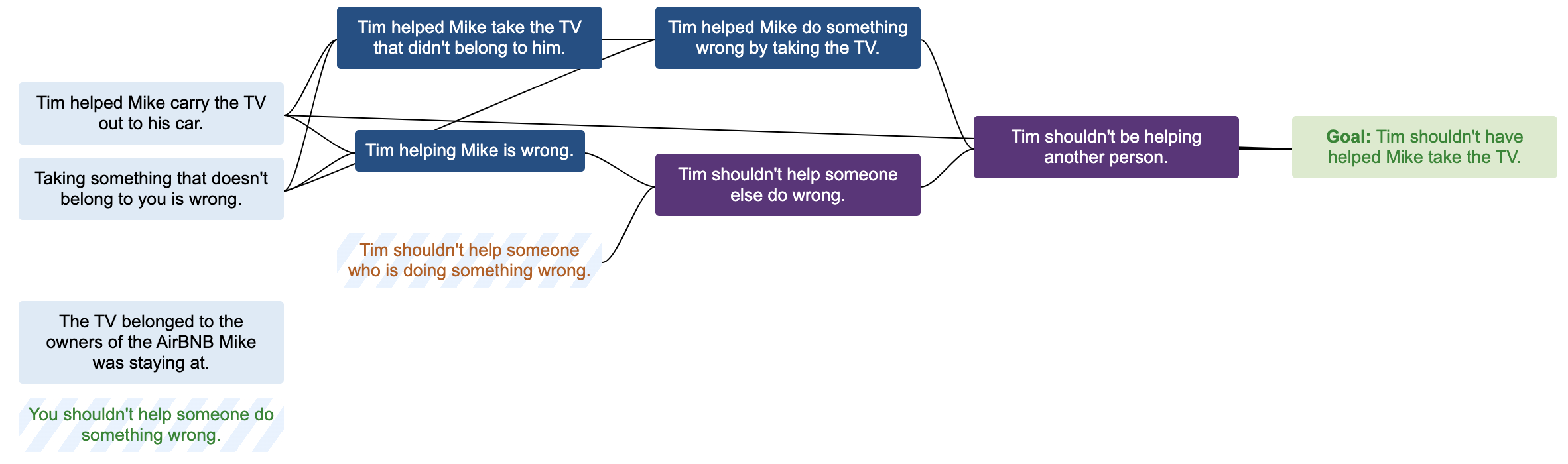}
    \includegraphics[width=1.0\linewidth]{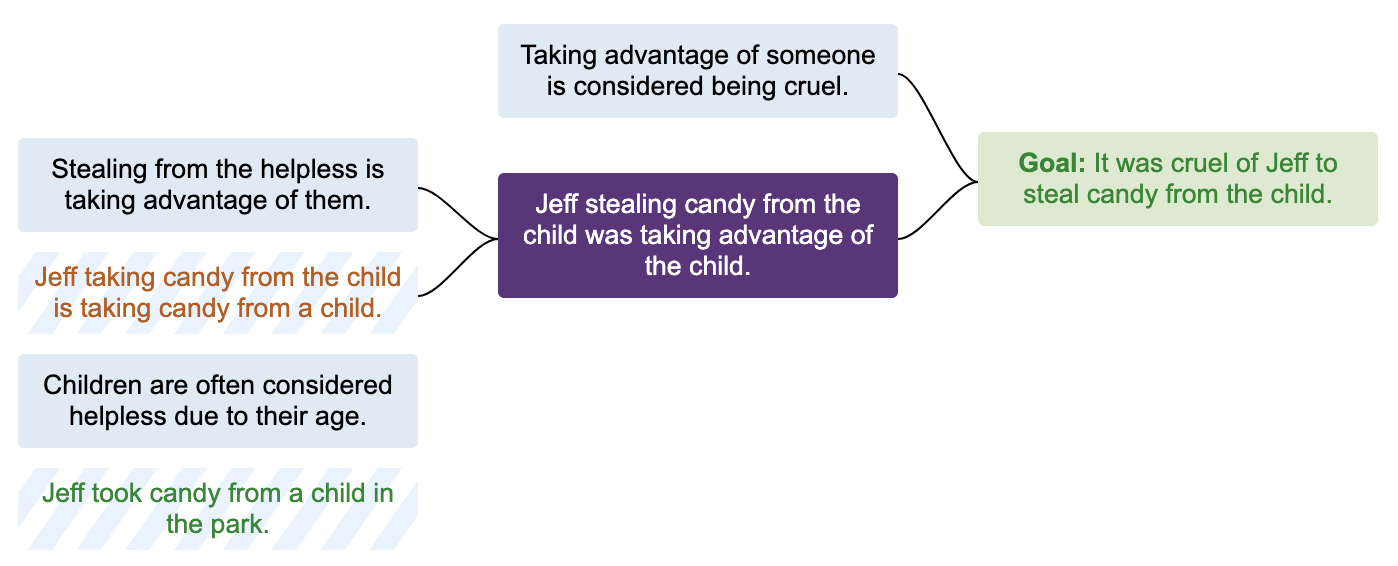}
    \includegraphics[width=1.0\linewidth]{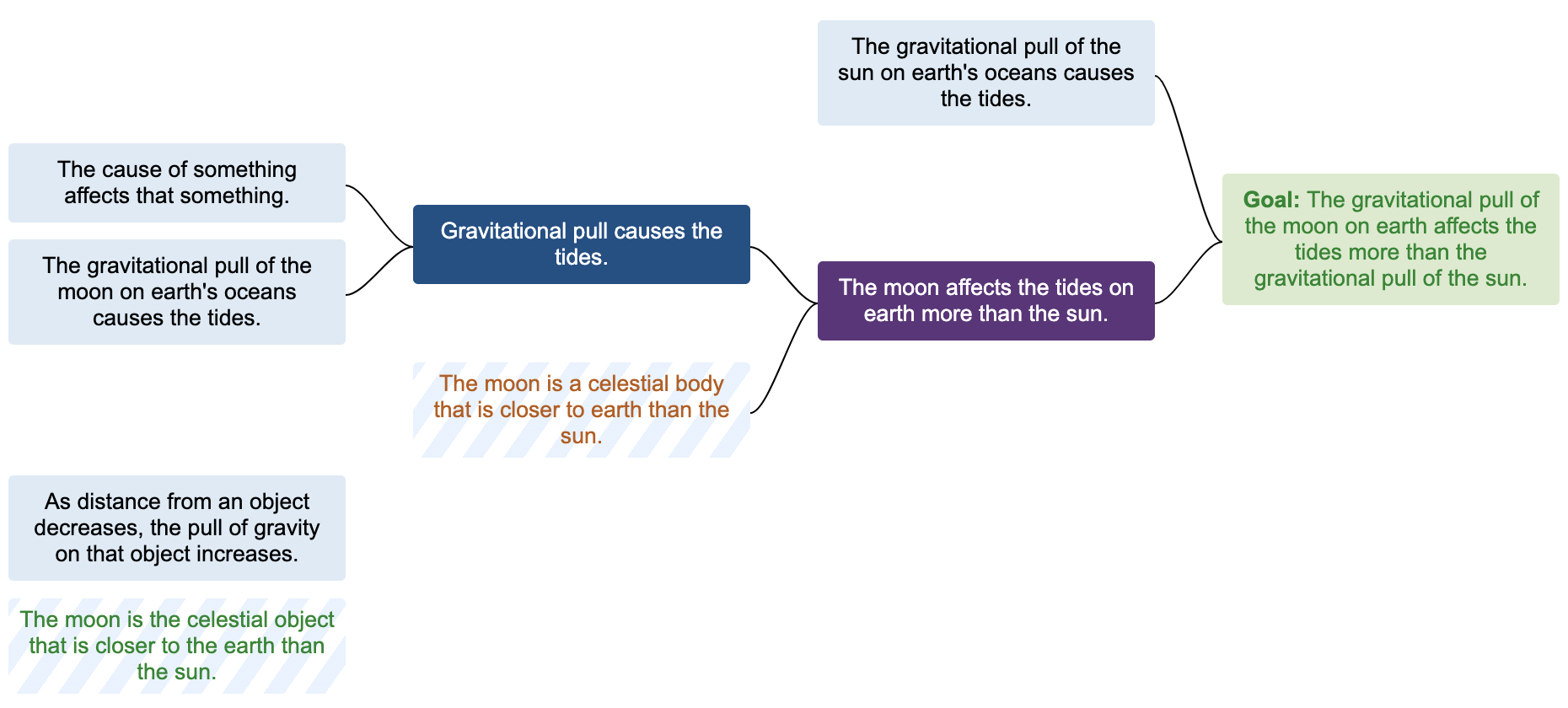}
    \caption{Example \textbf{successful proofs} using ADG from the Depth = all experiment. Boxes with blue stripes and orange text $\mathbf{x'}$ are generated premises from the abductive model where blue stripes with green text are the gold missing premise $\mathbf{x}_m$.  Light blue boxes are premises, dark blue are intermediate, purple are abductions, and green is the goal of the entailment tree. Note that the gold missing premise is never incorporated in the proof because we are trying to regenerate it through our step models.}
    \label{fig:adg_example_proofs}
\end{figure*}

We include multiple example proofs generated by our models from Table \ref{tab:main_metrics} at depth = all.  Each figure visualizes proofs from a specific model (ADGV on Figure \ref{fig:adgv_example_proofs} and ADG on Figure \ref{fig:adg_example_proofs}) and shows two examples from the ENWN dataset on the top and one example from the EntailmentBank dataset on the bottom. Furthermore, we show 8 examples of where the ADGV model failed to produce a proof with a caption explaining where the errors occurred.  The first 7 failed search examples are from ENWN and the last is from EntailmentBank.

\clearpage 
\begin{figure*}[b]
    \centering
    \vspace*{0cm}
    
    \includegraphics[width=1.0\linewidth]{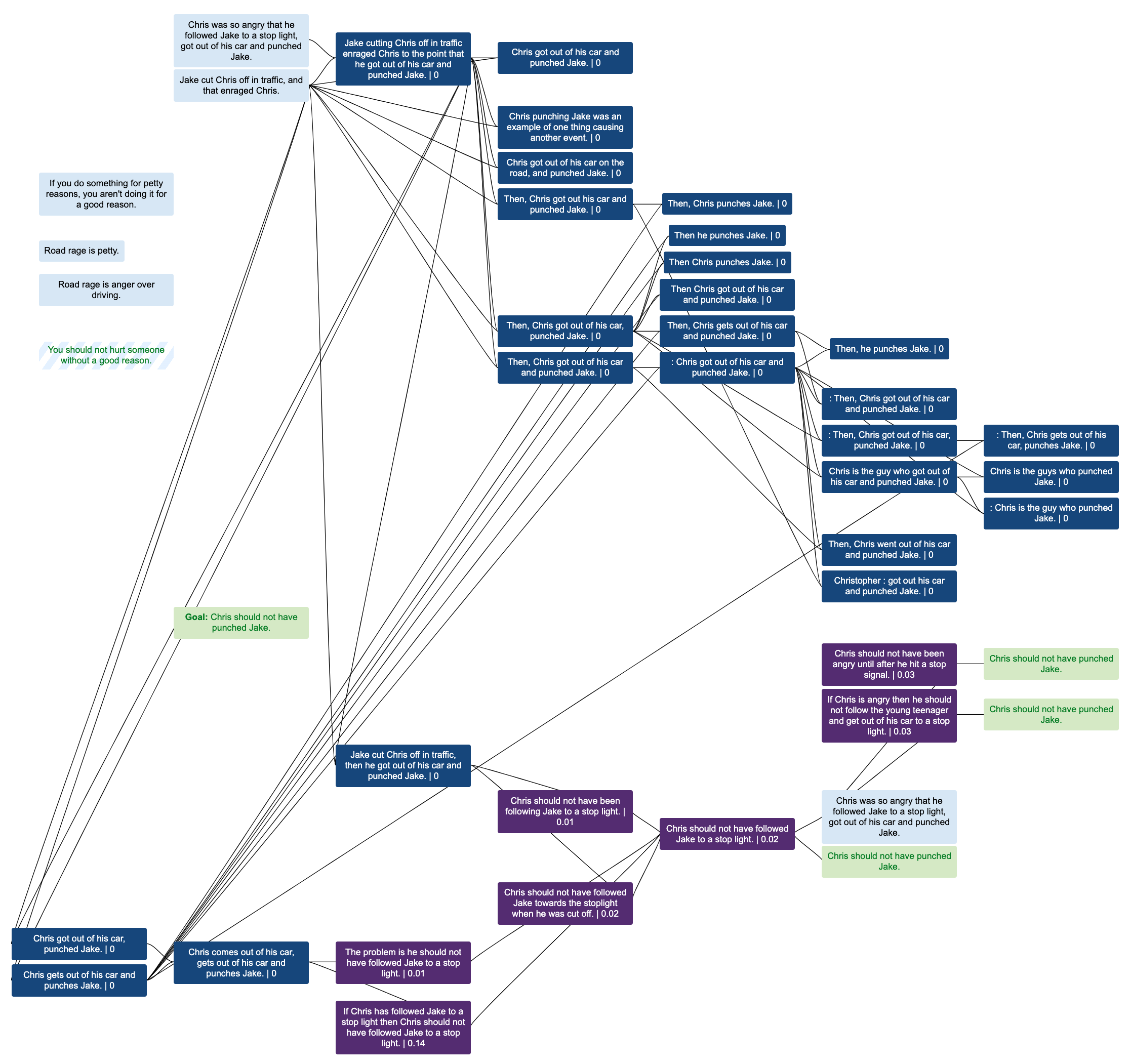}

    \caption{Example of a \textbf{failed search} using ADGV on the depth = all experiment for an example of the ENWN dataset.  Here the ADGV model fails to make use of all the premises given and continuously combines generations from a subset of the premises and their generations keeping the proof at a specific level of depth that's incapable of recovering the missing premise.}
    \label{fig:adgv_failed_example_1}
\end{figure*}

\begin{figure*}[b]
    \centering
    \vspace*{0cm}
    
    \includegraphics[width=1.0\linewidth]{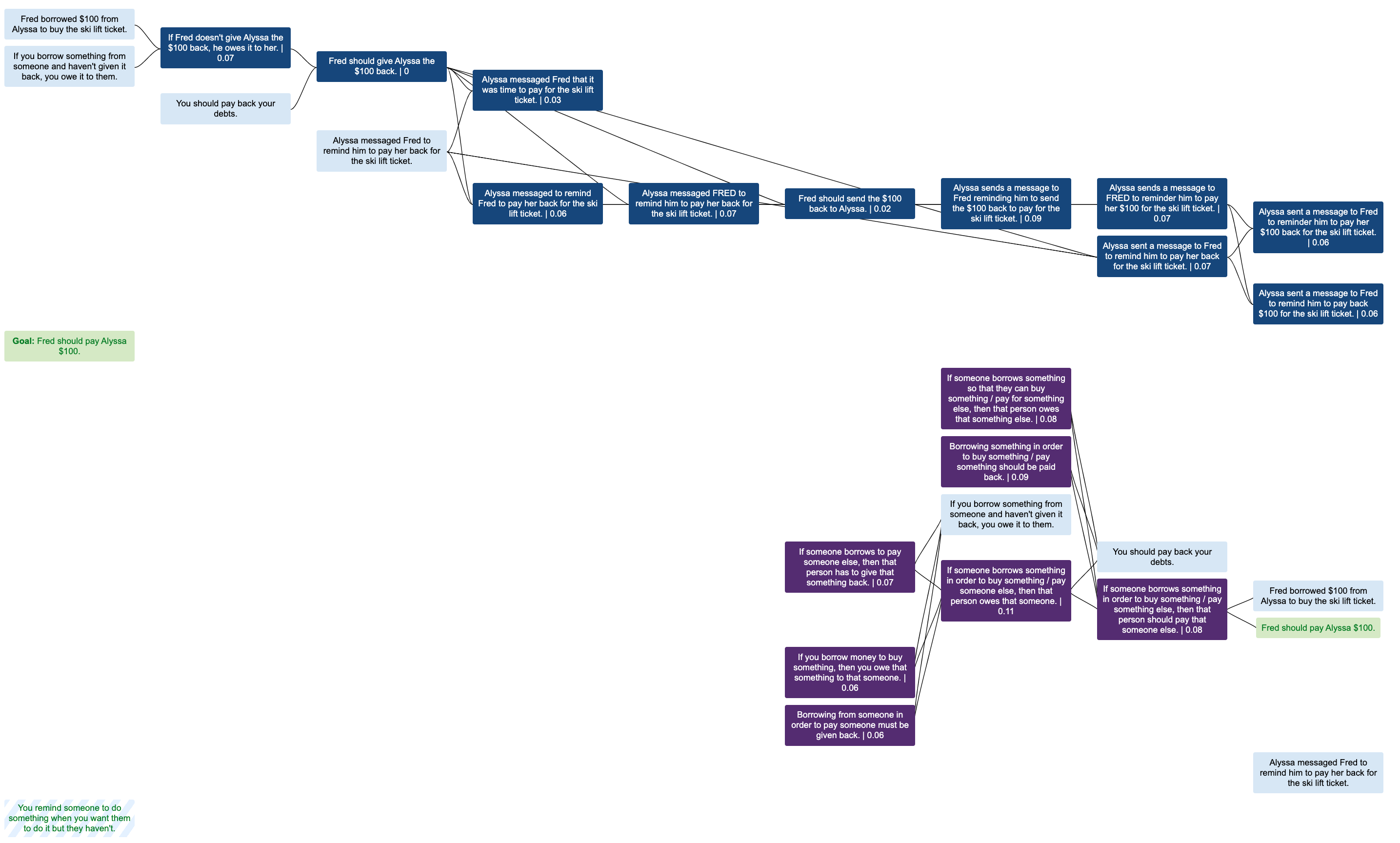}

    \caption{Example of a \textbf{failed search} using ADGV on the depth = all experiment for an example of the ENWN dataset. Here the two fringes, $f_d$ and $f_a$, are kept separate despite there being information beneficial for generating the missing premise $\mathbf{x}_m$ in the forward subtree.  We rank all abductive step combinations through the heuristic function $H_a$, which can fail to recognize useful combinations of forward deductions and open hypotheses. Ideally one of the deductions about Alyssa messaging Fred combined with an abductive generation about borrowing may have pushed the search further towards the missing premise, \emph{You remind someone to do something when you want them to do it but they haven't}.}
    \label{fig:adgv_failed_example_2}
\end{figure*}

\begin{figure*}[b]
    \centering
    \vspace*{0cm}
    
    \includegraphics[width=1.0\linewidth]{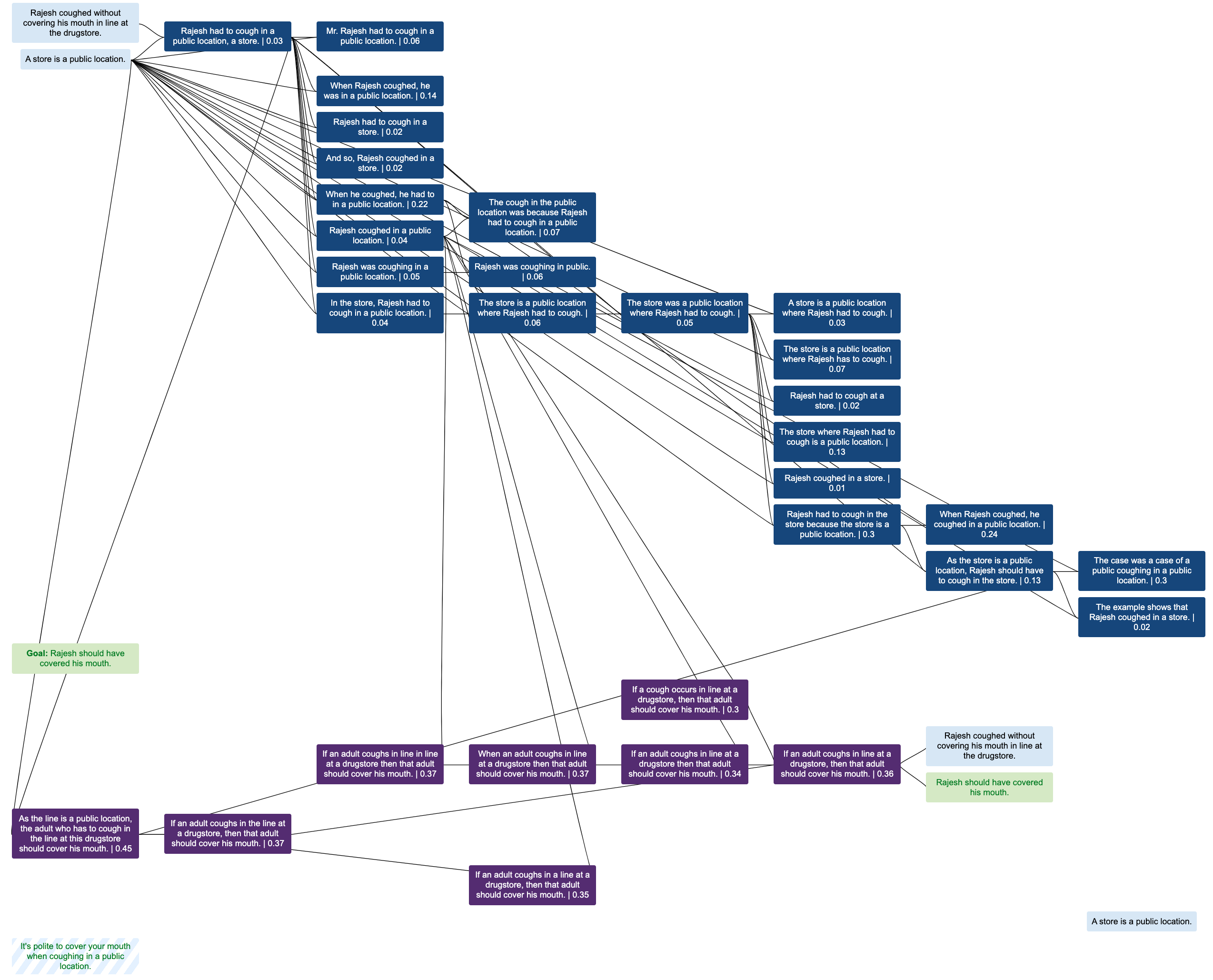}

    \caption{Example of a \textbf{failed search} using ADGV on the depth = all experiment for an example of the ENWN dataset.  This is an example of a difficult entailment tree to solve for step models.  The goal is to generate \emph{It's polite to cover your mouth when coughing in a public location.}, but because there is no premise that states \emph{It's impolite to do things you shouldn't do in public} (or something similar) the abductive model restrains itself in hallucinating such generations.  This leads to a close generation \emph{As the line is in a public location, the adult who has to cough in the line at this drugstore should cover his mouth}.  Although this isn't as general as the original missing premise, this generation is fairly close to it. Depending on how lenient the system is allowed to be this could be considered a false negative and is an example of the scoring metric being too restrictive.}
    \label{fig:adgv_failed_example_3}
\end{figure*}

\begin{figure*}[b]
    \centering
    \vspace*{0cm}
    
    \includegraphics[width=1.0\linewidth]{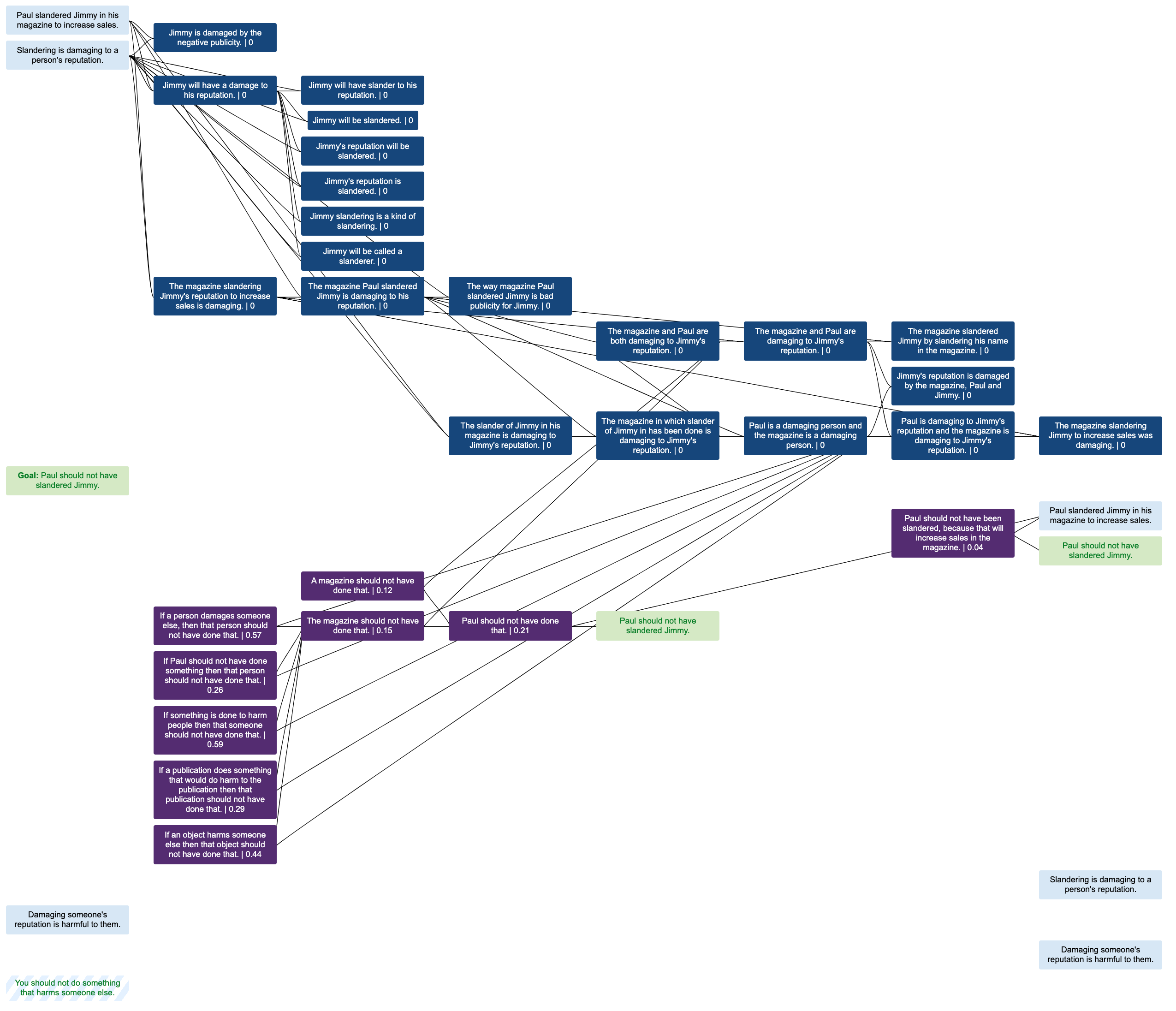}

    \caption{Example of a \textbf{failed search} using ADGV on the depth = all experiment for an example of the ENWN dataset.    This is an example of a potential false negative. In this example, the abductive model generates \emph{If a person damages someone else, then that person should not have done that.} which is extremely close (and semantically similar) to the gold missing premise \emph{You should not do something that harms someone else.} However, the entailment model scored the generation as 0.57 which is below our threshold for entailment.}
    \label{fig:adgv_failed_example_4}
\end{figure*}

\begin{figure*}[b]
    \centering
    \vspace*{0cm}
    
    \includegraphics[width=1.0\linewidth]{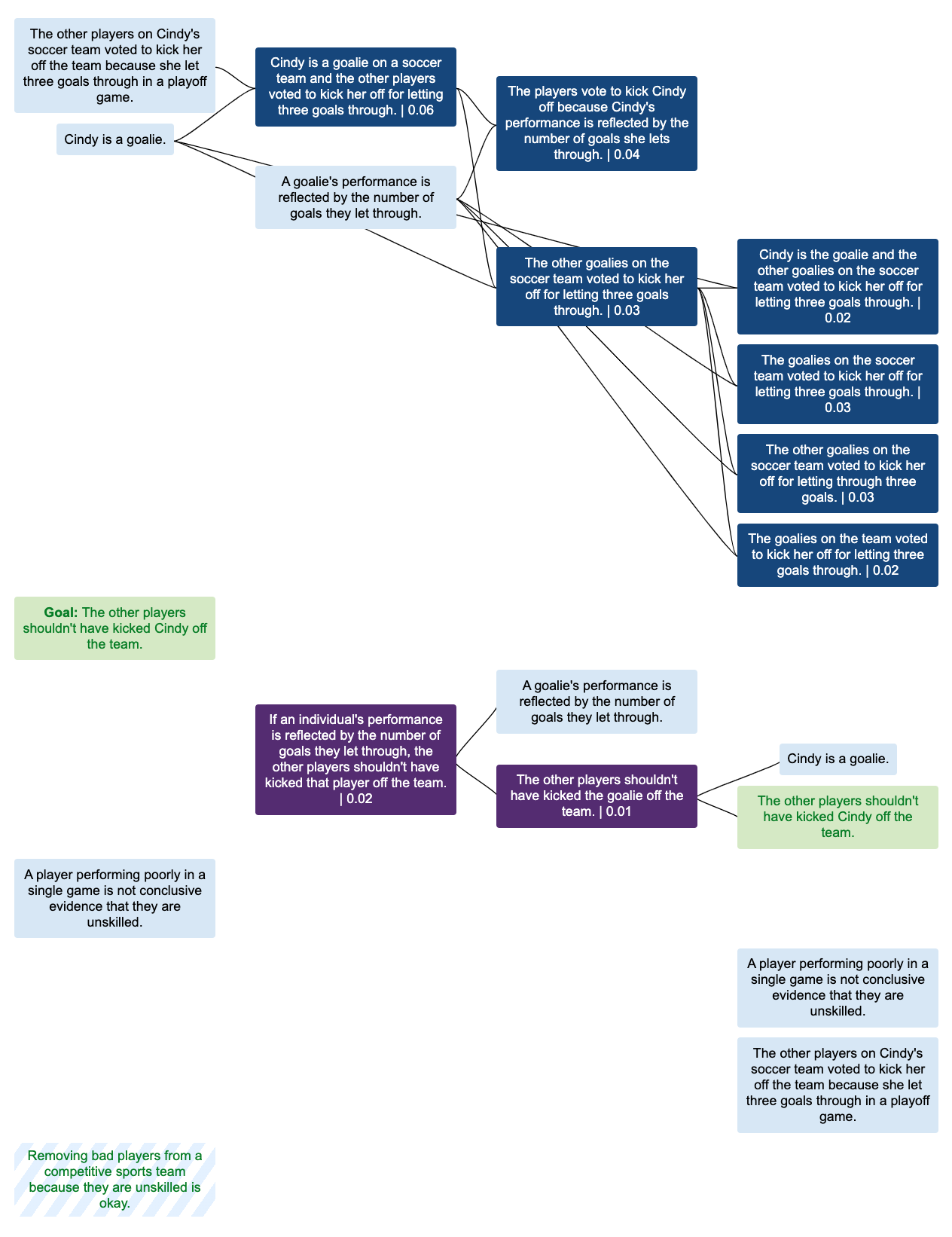}

    \caption{Example of a \textbf{failed search} using ADGV on the depth = all experiment for an example of the ENWN dataset.  Most of this search's generations were filtered out by one of the validators.  The second abductive generation, \emph{If an individual's...}, is an example of the abductive model following a similar pattern seen in EntailmentBank (the dataset used to train the model).  The abductive generation, although technically valid, is incorrect given the goal of the proof.  Instead, we would have wanted the abductive model to take the premise \emph{The other players on Cindy's soccer team voted to kick her off the team because she let three goals through in a playoff game.} and combine it with the first abductive generation \emph{The other players shouldn't have kicked the goalie off the team}. This is either a failure of the heuristic $H_a$ or a failure of the validators $V$ for removing the generations $y_a$ that came from that step. }
    \label{fig:adgv_failed_example_5}
\end{figure*}

\begin{figure*}[b]
    \centering
    \vspace*{0cm}
    
    \includegraphics[width=1.0\linewidth]{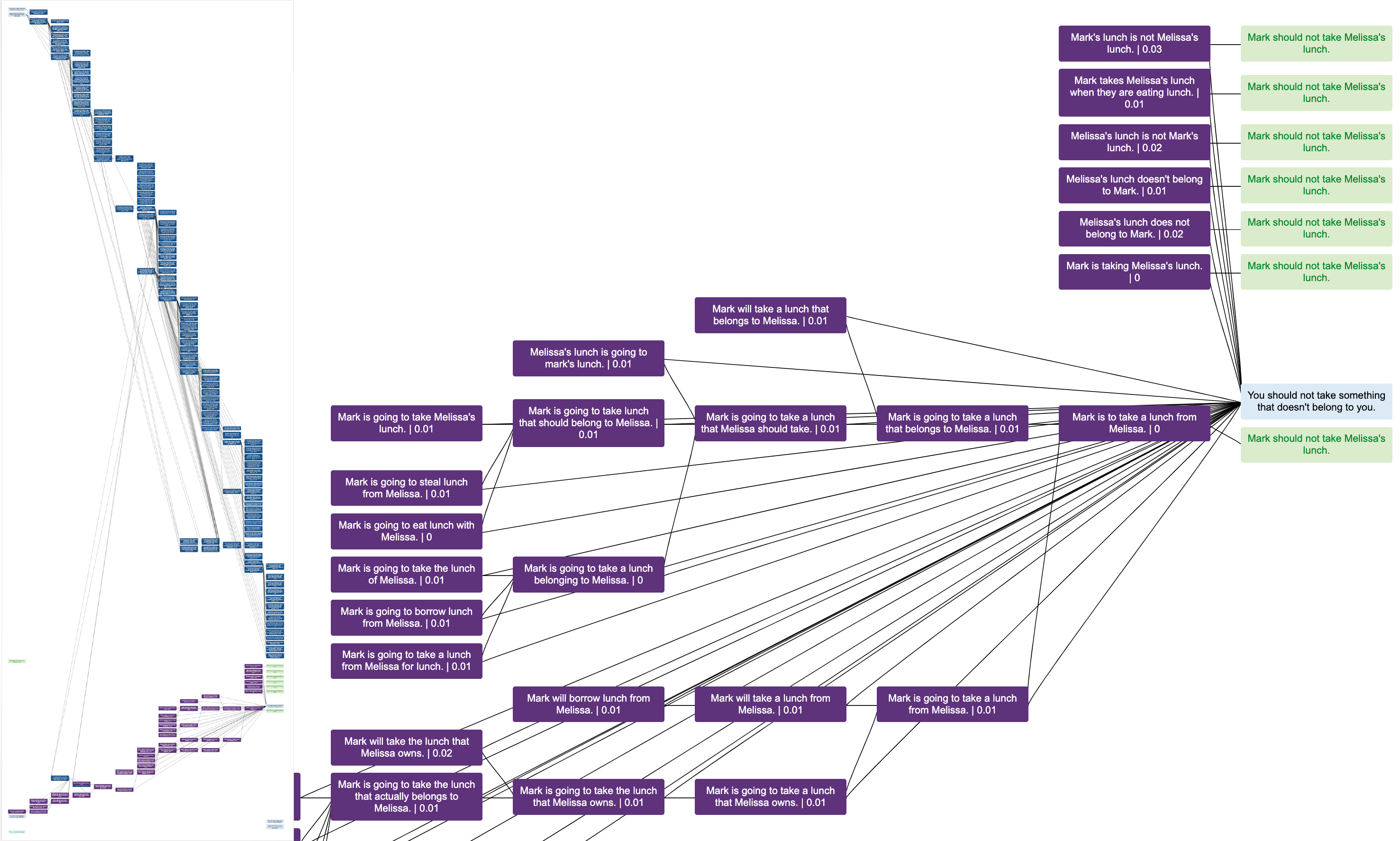}

    \caption{Example of a \textbf{failed search} using ADGV on the depth = all experiment for an example of the ENWN dataset. Although in the other failure cases we've shown the tree is somewhat small, most failure cases have an extremely large tree similar to the one in this figure on the right.  One of the more common failure modes is the recombination of premises and generations to create deeper proofs that restate the same information slightly differently (left part of this figure).  We tried to address this with the Consanguinity filter. Empirically we found that these slight tweaks in generations can lead to improved step recall, however, due to the scoring of both the entailment model and the heuristic functions being more favorable to specific phrasings.}
    \label{fig:adgv_failed_example_6}
\end{figure*}

\begin{figure*}[b]
    \centering
    \vspace*{0cm}
    
    \includegraphics[width=1.0\linewidth]{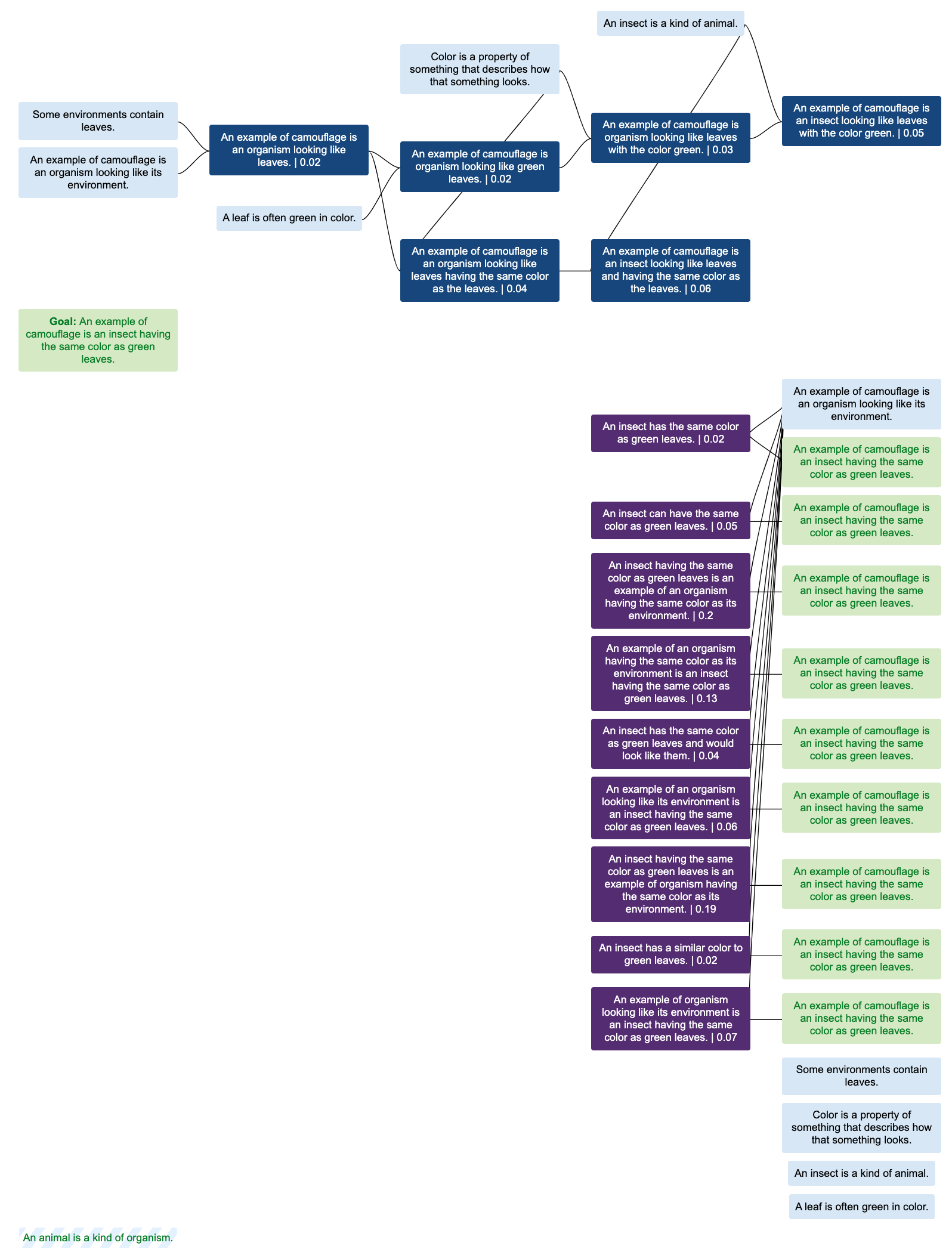}

    \caption{Examples of \textbf{failed searches} using ADGV on the depth = all experiment for an example of the EntailmentBank dataset. In this example the deductive fringe $f_d$ makes use of all the given premises but the abductive model does not.  The two fringes do not combine either.}
    \label{fig:adgv_failed_example_7}
\end{figure*}

\begin{figure*}[b]
    \centering
    \vspace*{0cm}
    
    \includegraphics[width=1.0\linewidth]{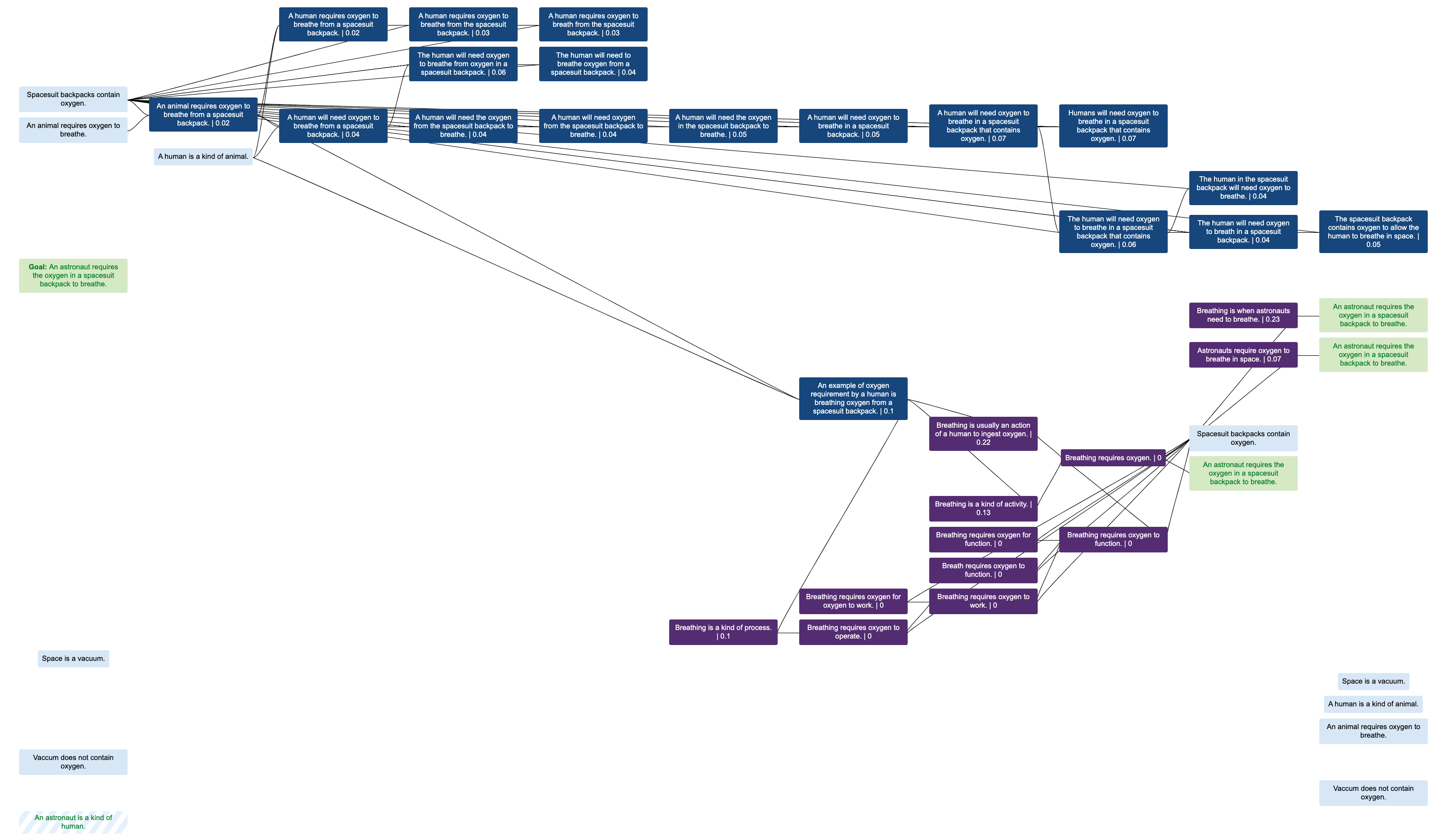}

    \caption{Example of a \textbf{failed search} using ADGV on the depth = all experiment for an example of the EntailmentBank dataset. Another example of the abductive fringe, $f_a$, not using all of its premises where \emph{An animal requires oxygen to breathe.} is paramount to recovering the missing premise.}
    \label{fig:adgv_failed_example_8}
\end{figure*}

\end{document}